\documentclass[runningheads]{llncs}
\pdfoutput=1 

\usepackage{eccv}

\usepackage{eccvabbrv}

\usepackage{graphicx}
\usepackage{booktabs}
\usepackage{amsmath}
\usepackage{amssymb}
\usepackage[table]{xcolor}
\usepackage{multirow}
\usepackage{array}
\usepackage{subcaption}
\usepackage{enumitem}
\usepackage{placeins}
\usepackage{makecell}

\usepackage[accsupp]{axessibility}  

\newcolumntype{C}{>{\centering\arraybackslash}p{1.05cm}}

\definecolor{lightred}{RGB}{253,191,191}
\definecolor{lightorange}{RGB}{255,223,191}
\definecolor{lightyellow}{RGB}{254,240,198}

\usepackage{hyperref}

\usepackage[capitalize]{cleveref}
\crefname{section}{Sec.}{Secs.}
\crefname{table}{Table}{Tables}
\crefname{figure}{Fig.}{Figs.}

\usepackage{orcidlink}

\begin{document}

\title{RL-AWB: Deep Reinforcement Learning for Auto White Balance Correction in Low-Light Nighttime Scenes}

\titlerunning{RL-AWB}

\author{Yuan-Kang Lee\inst{1,3*} \and Kuan-Lin Chen\inst{2*} \and Chia-Che Chang\inst{1} \and Yu-Lun Liu\inst{3}}

\authorrunning{Y.-K.~Lee et al.}

\institute{\textsuperscript{\rm 1} MediaTek Inc., Taiwan \quad \textsuperscript{\rm 2} National Taiwan University, Taiwan \\\textsuperscript{\rm 3} National Yang Ming Chiao Tung University, Taiwan\\
\email{r12942062@ntu.edu.tw, yulunliu@cs.nycu.edu.tw}\\
}

\renewcommand{\thefootnote}{\fnsymbol{footnote}}
\footnotetext[1]{The first two authors contributed equally to this work.}

\maketitle

\vspace{-1em}
\begin{figure}[htbp]
    \centering
    \includegraphics[width=\linewidth]{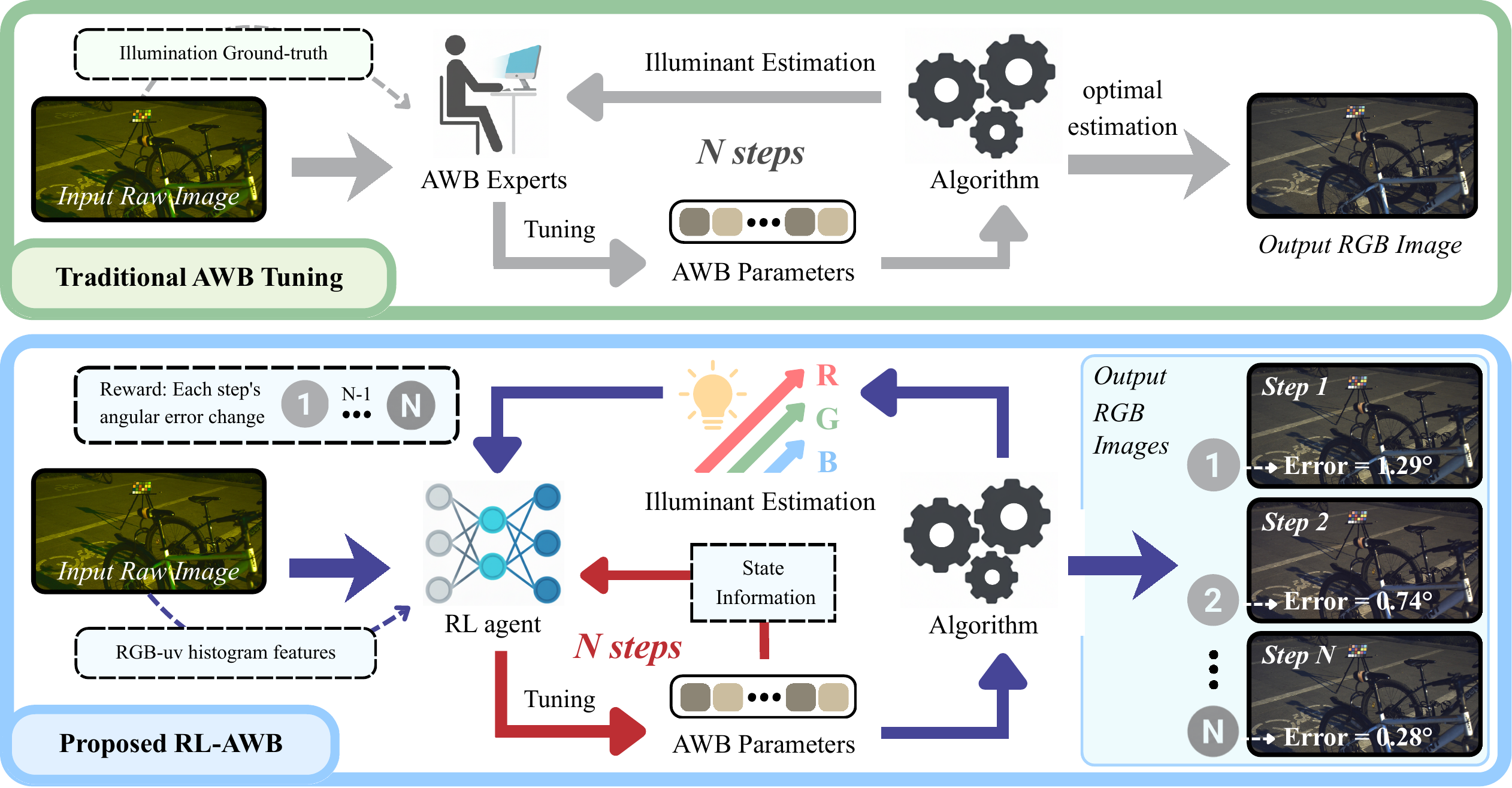}
    \caption{\textbf{Our method achieves optimal parameter tuning for automatic white balance (AWB) in nighttime scenes through a hybrid combination of a novel statistical color constancy algorithm and reinforcement learning.} Due to the complex lighting conditions in night scenes and the presence of significant noise in the images, traditional AWB tuning faces relatively difficult and time-consuming challenges when adapting parameters for night scene images. Our method can optimize parameters for different night scene images at a faster speed without requiring prior knowledge of illumination ground-truth, and has better cross-sensor generalization advantages.}
    \label{fig:RLAWBteaser}
\end{figure}
\vspace{-2em}

\begin{abstract}
Nighttime color constancy still remains a challenging problem in computational photography due to low-light noise and complex illumination conditions. We present RL-AWB, a novel framework combining statistical methods with deep reinforcement learning for nighttime white balance. Our method begins with a statistical algorithm tailored for nighttime scenes, integrating salient gray pixel detection with novel illuminant estimation. Building on this foundation, we develop the first deep reinforcement learning approach for color constancy that leverages the statistical algorithm as its core, mimicking professional AWB tuning experts by dynamically determining image-specific parameters at inference time, without requiring ground-truth illuminants or reference images. To further facilitate cross-sensor evaluation, we introduce the first multi-sensor nighttime dataset. Experiment results demonstrate that our method achieves strong generalization capability across low-light and well-illuminated images. Project page: \url{https://ntuneillee.github.io/research/rl-awb/}

\keywords{Auto White Balance \and Color Constancy \and Reinforcement Learning \and Nighttime Imaging \and Low-light}
\end{abstract}

\section{Introduction}
\label{sec:intro}

Auto White Balance (AWB) is a fundamental component of camera image signal processing (ISP) pipelines that estimates scene illuminant and corrects color casts, ensuring white objects appear neutral across varying lighting conditions~\cite{buchsbaum1980spatial,finlayson2004shades}. While existing methods achieve robust performance in well-lit daytime scenarios~\cite{barron2015convolutional,afifi2021cross,chang2025gcc}, nighttime environments present a fundamentally different challenge. Low illumination, high ISO settings, and severe chromatic noise violate the statistical assumptions underlying conventional AWB algorithms~\cite{van2007edge,qian2019finding}, leading to highly unstable illuminant estimates. This instability is further exacerbated in cross-sensor deployment, where the same model often produces significant and unpredictable color shifts across different camera sensors and ISP configurations~\cite{cheng2014illuminant,afifi2021cross}. Addressing robust cross-sensor nighttime AWB is therefore critical for real-world applications, including mobile photography, surveillance systems, and automotive imaging.

The core difficulty of nighttime AWB stems from the breakdown of reliable color statistics. Statistical methods~\cite{buchsbaum1980spatial,finlayson2004shades,van2007edge} assume sufficient scene diversity and stable gray pixel detection, which fail under extreme low-light conditions where sensor noise dominates signal~\cite{cheng2024nighttime}. Deep learning approaches~\cite{barron2015convolutional,hu2017fc4,chang2025gcc}, while effective in daytime scenarios, require extensive labeled nighttime training data and suffer from severe generalization degradation when deployed on unseen camera sensors~\cite{afifi2021cross,kim2025ccmnet}. Furthermore, nighttime scenes exhibit heightened sensitivity to algorithmic parameters: small variations in parameter selection can lead to dramatically different illuminant estimates.

To address these challenges, we present {RL-AWB}, the \textit{first} framework that integrates reinforcement learning into automatic white balance for nighttime color constancy. Our approach fundamentally differs from existing paradigms by formulating AWB as a sequential decision-making problem, where an RL agent learns adaptive parameter selection policies for a novel statistical illuminant estimator. This hybrid design preserves the interpretability and sensor-agnostic nature of statistical methods~\cite{finlayson2004shades,qian2019finding} while gaining the adaptive capability of learning-based approaches~\cite{barron2015convolutional,afifi2021cross}, all with minimal training data requirements.

In summary, we make the following contributions:
\begin{itemize}
\item We develop {SGP-LRD} (Salient Gray Pixels with Local Reflectance Differences), a nighttime-specific color constancy algorithm that achieves state-of-the-art illuminant estimation on public nighttime benchmarks.
\item We design the {RL-AWB framework} with Soft Actor-Critic training and two-stage curriculum learning, enabling adaptive per-image parameter optimization with exceptional data efficiency.
\item We contribute {LEVI}, the first multi-camera nighttime dataset comprising 700 images, enabling rigorous cross-sensor color constancy evaluation.
\item With only 5 training images per dataset, RL-AWB generalizes across unseen sensors better than fully-supervised state-of-the-art methods, demonstrating high data efficiency.
\end{itemize}

\begin{figure}[tp]
    \centering
    \includegraphics[width=\linewidth]{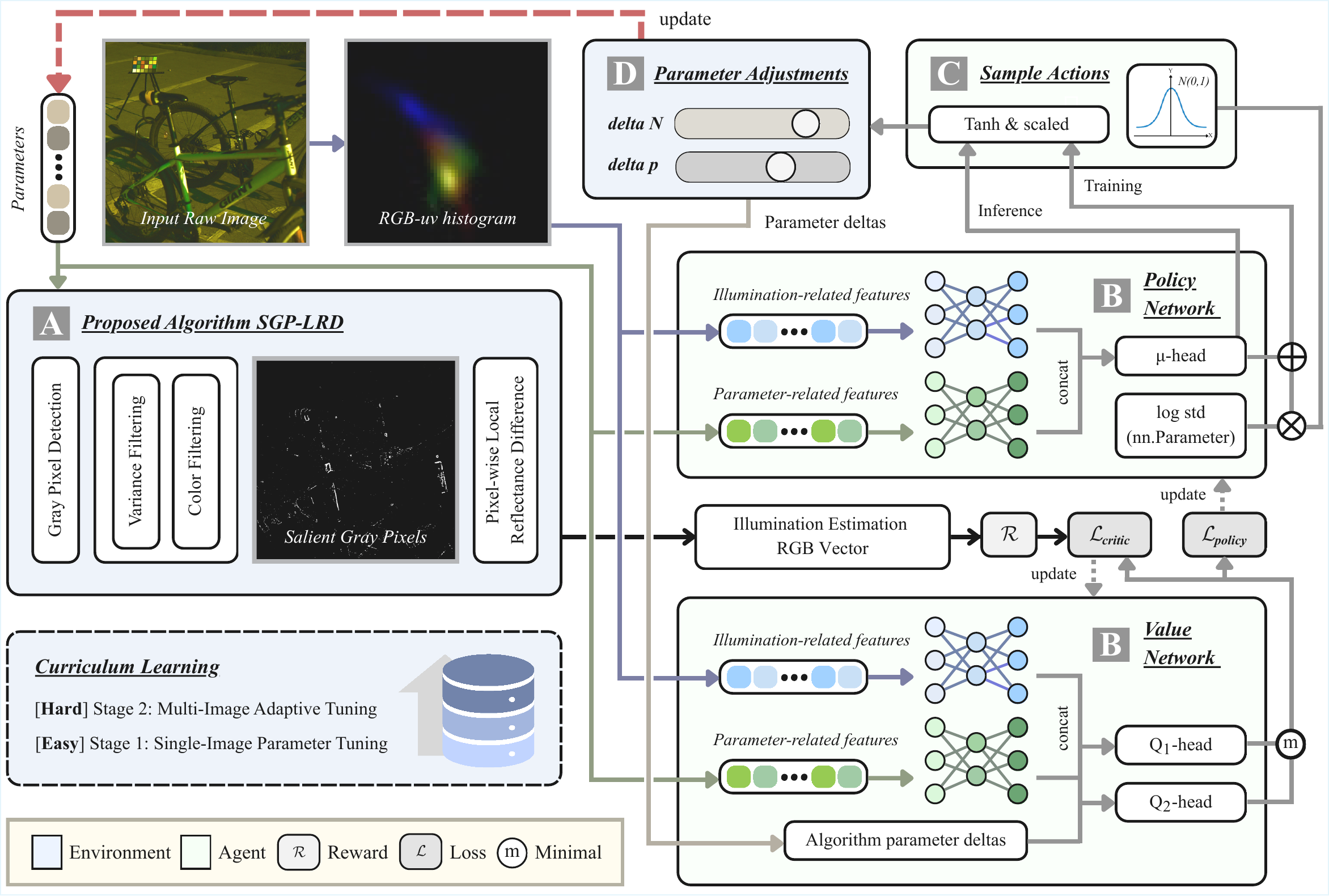}
    \caption{\textbf{Overview of the proposed RL-AWB framework.}
    (A) Given an input image, the proposed nighttime color constancy algorithm SGP-LRD estimates the scene illuminant conditioned on two hyperparameters (gray-pixel sampling percentage $N$ and Minkowski order $p$).
    (B) A SAC agent selects parameter updates based on image statistics and current AWB settings.
    (C) The policy outputs one action per parameter; actions are sampled, squashed by $\tanh$ to $[-1,1]$, and rescaled to valid ranges.
    (D) The rescaled actions update the two hyperparameters and are applied to SGP-LRD to produce the illuminant estimate. Repeat until the termination criterion is met.}
    \label{fig:flowchart}
\end{figure}

\section{Related Work}
\label{sec:related}

\paragraph{Statistical Color Constancy.}
Statistical methods exploit scene statistics via achromatic averages (Gray World~\cite{buchsbaum1980spatial}), maximum responses (Max-RGB) rooted in Retinex theory~\cite{land1971lightness}, unified by Minkowski norms~\cite{finlayson2004shades}. Gamut mapping methods~\cite{forsyth1990novel,finlayson2000improving,gijsenij2010generalized} constrain feasible illuminants using observed colors, while Bayesian approaches~\cite{brainard1997bayesian,gehler2008bayesian} formalize probabilistic priors over illuminant distributions. Edge-based approaches~\cite{van2007edge} use derivatives with photometric weighting~\cite{gijsenij2011improving} and local statistics~\cite{ebner2009color}. Corrected-moment estimation~\cite{finlayson2013corrected} and spatio-spectral statistics~\cite{chakrabarti2011color} further improve classical estimators. Recent advances include grayness indices~\cite{qian2019finding}, FFT acceleration (700 FPS)~\cite{barron2017fast}, spectral methods~\cite{koskinen2024single}, attention mechanisms~\cite{laakom2020bag}, and multi-scale spatial statistics~\cite{ulucan2024multi}. For comprehensive evaluations, we refer readers to~\cite{gijsenij2011computational,barnard2002comparison}. \emph{Unlike} these methods using \emph{fixed} parameters or heuristics, we employ reinforcement learning to \emph{automatically} learn optimal parameter adjustments for varying nighttime conditions, combining statistical efficiency with adaptive optimization.

\paragraph{Learning-Based Color Constancy.}
Deep learning approaches feature CNNs with log-chrominance localization and confidence pooling~\cite{barron2015convolutional,hu2017fc4}, enhanced by cascades~\cite{yu2020cascading}, contrastive learning~\cite{lo2021clcc}, and adaptation~\cite{afifi2021cross}. Earlier works include CNN-based illuminant classification~\cite{bianco2015color,oh2017approaching}, multi-illuminant CNN estimation~\cite{bianco2017single}, deep specialized networks with multi-hypothesis selection~\cite{shi2016deep}, and metric learning formulations~\cite{xu2020end}. Multi-hypothesis approaches~\cite{hernandez2020multi} further enable camera-agnostic probabilistic estimation. GAN-based methods~\cite{sidorov2019conditional,das2021generative} address multi-illuminant scenarios through domain translation. Quasi-unsupervised training~\cite{bianco2019quasi} and few-shot meta-learning~\cite{mcdonagh2018formulating} reduce label requirements. Recent diffusion methods~\cite{chang2025gcc} achieve 4.32°--5.22° worst-25\% error via inpainting, while self-supervised~\cite{peng2024cccg} and uncertainty-aware~\cite{buzzelli2025uncertainty} approaches further reduce supervision. Extensions address multi-illuminant~\cite{li2022transcc}, post-editing~\cite{afifi2020deep}, mixed lighting~\cite{afifi2022auto}, style-based AWB~\cite{kinli2023modeling}, and brightness~\cite{xie2025boosting,chen2023learning}. Simple features with ensemble learning~\cite{cheng2015effective} bridge statistical and deep learning paradigms. \emph{In contrast to} these \emph{data-hungry} approaches requiring extensive labeled nighttime data, our method achieves cross-dataset generalization through \emph{sample-efficient} RL tuning of interpretable statistical parameters, combining both paradigms.

\paragraph{Nighttime and Low-Light Color Constancy.}
Nighttime scenes present unique challenges that fundamentally violate daytime color constancy assumptions: mixed illumination from multiple light sources, severely low light levels, and elevated sensor chroma noise~\cite{chang2020learning}. Early methods~\cite{chen2018learning,wei2018deep,guo2020zero} targeted brightness. Recent work employs transformers~\cite{cai2023retinexformer,xu2022snr,wang2023ultra}, adaptive masking~\cite{li2024nightcc}, synthetic data~\cite{punnappurath2022day}, noise-resistant detection~\cite{cheng2024nighttime}, and joint restoration~\cite{zhou2022lednet,ma2022toward,yi2023diff,wu2023learning}. Diffusion-based approaches~\cite{jiang2023low,wang2023exposurediffusion} and Fourier-domain methods~\cite{li2023embedding,wang2023fourllie} offer new paradigms for low-light enhancement, while novel color spaces~\cite{yan2025hvi} address color handling under extreme darkness. Nighttime rendering benchmarks~\cite{liu2024ntire,li2022rendering} further establish standardized evaluation protocols. For a comprehensive survey, see~\cite{li2021low}. However, these methods rely on \emph{pseudo-labels} with error propagation or \emph{fixed} parameters. We formulate nighttime AWB as \emph{sequential decision-making}, where RL learns \emph{dynamic} parameter adjustment strategies.

\paragraph{Reinforcement Learning for ISP.}
RL enables adaptive ISP policies for exposure~\cite{yu2018deepexposure,hu2018exposure}, color enhancement~\cite{park2018distort}, rapid convergence~\cite{lee2024learning} (5 frames, 1ms), and artifact filtering~\cite{bajaj2023reinforcement}. The seminal RL-Restore~\cite{yu2018crafting} introduced toolchain selection for image restoration via DQN. Applications span pixel-wise correction~\cite{furuta2019pixelrl}, software parameters~\cite{kosugi2020unpaired}, personalization~\cite{yang2018personalized,zhao2025ref}, sequential optimization~\cite{sun2024rl}, and module selection~\cite{wang2024adaptiveisp,zhang2024efficient}. Our framework adopts the Soft Actor-Critic algorithm~\cite{haarnoja2018sac,haarnoja2018soft} for its sample-efficient off-policy updates and automatic entropy tuning. These works validate RL for ISP but focus on \emph{daytime} scenarios with \emph{camera settings} as actions. We extend to \emph{nighttime} AWB where actions control \emph{statistical algorithm parameters}, requiring noise-aware rewards and robust optimization.

\paragraph{Cross-Sensor Generalization.}
Cross-sensor generalization remains a challenge in color constancy, as different camera sensors exhibit varying spectral responses and ISP characteristics~\cite{liu2020single}. Methods evolved from dataset evaluation~\cite{cheng2014illuminant} through fine-tuning~\cite{afifi2021cross}, sensor-independent representations~\cite{afifi2019sensor}, embeddings~\cite{kim2025ccmnet}, to domain-invariant learning~\cite{zhang2022domain,tang2022transfer}. Multi-camera benchmarks~\cite{kim2021large} establish standardized cross-sensor evaluation. Calibration-light methods use dual-mapping~\cite{yue2024effective} (single D65), self-supervision~\cite{cun2022learning}, multi-domain architectures~\cite{xiao2020multi}, and HDR~\cite{afifi2024optimizing}. Test-time adaptation~\cite{wang2020tent,liu2021ttt++,sun2020test,karmanov2024efficient} enables training-free deployment, with recent advances in continual~\cite{wang2022continual} and improved~\cite{chen2023improved} test-time strategies. \emph{Unlike} methods requiring \emph{test-time} multi-image access or \emph{camera-specific} calibration, we achieve generalization through (1) \emph{sensor-agnostic} statistical algorithms and (2) curriculum learning exposing agents to diverse conditions, enabling \emph{few-shot} inference on unseen cameras.

\paragraph{Hybrid Statistical-Learning Approaches.}
Hybrid methods combine classical algorithms with deep learning. Algorithm unrolling, pioneered by LISTA~\cite{gregor2010learning} and surveyed in~\cite{monga2021algorithm}, transforms iterative optimization into trainable networks, as demonstrated for image restoration~\cite{zhang2018ista,zhang2020deep,mou2022deep}. Differentiable ISP pipelines~\cite{mosleh2020hardware,tseng2019hyperparameter,yu2021reconfigisp} enable end-to-end optimization of classical processing modules. Other hybrid strategies include learned parameter dictionaries~\cite{conde2022model}, energy landscapes~\cite{bai2017deep}, and hyperparameter prediction~\cite{li2023learning}. Our work follows this paradigm, combining \emph{statistically-grounded} illuminant estimation with \emph{RL-based} parameter optimization and preserving interpretability.

\paragraph{Curriculum Learning in RL.}
Curriculum learning, introduced by Bengio~\emph{et al.}~\cite{bengio2009curriculum} and extended via self-paced learning~\cite{kumar2010self,jiang2015self}, improves training efficiency through progressive difficulty. In the RL setting, survey works~\cite{narvekar2020curriculum,soviany2022curriculum} systematize task sequencing strategies, including teacher--student approaches~\cite{matiisen2019teacher}, reverse curriculum generation~\cite{florensa2017reverse,tao2024reverse}, self-paced deep RL~\cite{klink2020self}, automated task selection~\cite{graves2017automated}, and strategic sampling~\cite{schaul2015prioritized,andrychowicz2017hindsight}. Our two-stage approach: (1) single-instance stabilization addresses cold starts, (2) cyclic multi-instance balances exploration and stability.

\section{Method}
\label{sec:method}

\subsection{Nighttime Color Constancy Algorithm}
\label{subsec:Nighttime Color Constancy Algorithms}

\paragraph{Salient Gray Pixel Detection.}
Under the narrow-band spectral assumption, the linear image $\mathbf{I}$ is modeled as $\mathbf{I} = \mathbf{W} \cdot \mathbf{L}+\delta$, where $\mathbf{W}$ is the white-balanced image, $\mathbf{L}$ is illuminant, and $\delta$ is noise. Neglecting $\delta$ and applying log-transform followed by local contrast operator $C\{\cdot\}$ (Laplacian of Gaussian) yields:
\begin{equation}
C\{ \log(I_i^{(x, y)}) \} \approx C\{ \log(W_i^{(x, y)}) \}.
\end{equation}
This shows local contrast depends solely on surface reflectance. Let $\Delta_i(x,y)$ denote the local contrast at $(x,y)$ for channel $i \in \{R,G,B\}$. Following Qian \textit{et al.}~\cite{Qian2019Revisiting}, achromatic pixels have contrast vectors $\Delta(x,y)=[\Delta_{R},\Delta_{G},\Delta_{B}]^{\mathrm{T}}$ aligned with the gray direction $\mathbf{g}=[1,1,1]^{\mathrm{T}}$. Grayness is measured as:
\begin{equation}
\label{eq:angular}
G(x, y) = \cos^{-1} \left( \frac{\langle \Delta_{(x, y)}, \mathbf{g} \rangle}{\|\Delta_{(x, y)}\| \|\mathbf{g}\|_2} \right).
\end{equation}
The top $N\%$ pixels ranked by $G(x,y)$ are selected as gray candidates. We then apply a two-layer filtering process to refine this candidate set, mitigating the adverse effects of noise and chromatic outliers prevalent in low-light imagery:

\paragraph{Local Variance Filtering.} For each pixel in the initial detected gray pixel mask, we compute the variance across the logarithmic RGB channels. Pixels where the intra-pixel variance is too small often lack reliable color information. By applying a lower bound threshold, $\mathrm{VarTh}$, we filter out these unreliable candidates.

\paragraph{Color Deviation Filtering.} The second stage filters out pixels that are too distant from the dominant color cast of the scene's illuminant. We first compute the mean logarithmic intensity for each channel of the image: $\mathbf{M} = [\bar{M}_R, \bar{M}_G, \bar{M}_B]^{\mathrm{T}}$, and then calculate the maximum absolute color deviation, $X(x,y)$, for each pixel from this mean. We define a threshold $T_C = \mathrm{ColorTh} \cdot \min(\mathbf{M})$. Any initial detected gray pixel exceeding this deviation is removed.

Following these two refinement layers to the initial detected gray pixels, we obtain the Salient Gray Pixels (SGPs).

\paragraph{Gray-pixel Confidence Weighting.} Let $R(i,j)$, $G(i,j)$, and $B(i,j)$ denote the normalized pixel intensities at location $(i,j)$ in the three channels, respectively. The luminance map for SGPs is then computed as $LM(i,j) = (R(i,j) + G(i,j) + B(i,j))/3$. The skewness value $s_{LM}$ quantifies the asymmetry of the brightness distribution and guides the selection of an adaptive exponent parameter $E$. We set $E = 1.0$ for highly skewed distributions ($s_{LM} > 1.5$), $E = 2.0$ for moderate skewness ($0.2 < s_{LM} \leq 1.5$), and $E = 4.0$ for uniform illumination ($s_{LM} \leq 0.2$). Let $\overline{LM}$ denote the mean luminance of non-zero SGPs. The gray-pixel confidence weight is then computed as:
\begin{equation}
W_{SGP}(i,j) = 1 - \exp\left[-\left(\frac{LM(i,j)}{\overline{LM}}\right)^E \ \right].
\end{equation}

\paragraph{Pixel-wise Local Reflectance Difference.} For each position $(i,j)$ in an image, a local window $\mathcal{W}_{i,j}$ of size $3 \times 3$ is centered at that pixel. Let $f_c(x)$ denote the detected SGP intensity for channel $c$, and $\mathcal{W}_{i,j}^*$ represent the set of non-zero pixels within $\mathcal{W}_{i,j}$. The pixel-wise normalized local reflectance difference at position $(i,j)$ is then defined as:
\begin{equation}
N_c^{i,j} = \frac{\left(\displaystyle\sum\limits_{x \in \mathcal{W}_{i,j}^*} f_c(x)\right) / \ |\mathcal{W}_{i,j}^*|}{\displaystyle\max\limits_{x \in \mathcal{W}_{i,j}} f_c(x)}.
\end{equation}
For pixels where the maximum value is zero, we set $N_c^{i,j} = 0$ to exclude invalid regions from the computation. Finally, the illuminant is estimated as:
\begin{equation}
\hat{e}_c = \left(\frac{\displaystyle\sum\nolimits_{\Omega} \left(\mu_c^{i,j} \cdot W_{SGP}(i,j)\right)^p}{\displaystyle\sum\nolimits_{\Omega} \left(N_c^{i,j} \cdot W_{SGP}(i,j)\right)^p}\right)^{1/p},
\end{equation}
where $\Omega$ represents all valid pixel positions ($N_c^{i,j} > 0$), $\mu$ is the mean intensity of SGPs in $W_{i,j}$, and $p$ is the Minkowski norm parameter.
The numerator accumulates weighted SGP intensity, and the denominator accumulates weighted normalized local differences.

Our proposed algorithm, SGP-LRD (Salient Gray Pixels with Local Reflectance Differences), addresses nighttime color constancy through three key design principles:

\begin{itemize}
    \item \textbf{Reliability amplification:} Spatially coherent gray regions are sampled repeatedly across overlapping windows, naturally amplifying high-SNR grayness signals while suppressing isolated spurious pixels.

    \item \textbf{Implicit noise filtering:} While spurious pixels from sensor noise appear in limited windows with minimal contribution, genuine gray regions exhibit consistent responses across neighboring windows, naturally distinguishing signal from noise through spatial redundancy.

    \item \textbf{Spatial prior exploitation:} The overlapping design encodes the natural prior that reliable achromatic surfaces exhibit spatial continuity, ensuring that illuminant estimates are dominated by high-confidence information.

\end{itemize}

\subsection{RL-AWB Framework}
\label{subsec:RL-AWB Framework}

\paragraph{Algorithm Parameters.} Two parameters critically determine the performance of our SGP-LRD: the gray pixel candidate selection threshold $N\%$ and the Minkowski norm exponent $p$. Lower $N\%$ for gray-rich scenes (higher purity); raise it when gray cues are sparse (better coverage). Small $p$ yields near-uniform weighting; large $p$ emphasizes high-confidence pixels; it is helpful when detections are reliable but brittle in ambiguous/low-light scenes. In low-light nighttime images, the optimal $(N, p)$ configuration is inherently scene-dependent. We address this challenge through a reinforcement learning framework that learns to adaptively select algorithm parameters based on scene characteristics.

\paragraph{State Design.}
Our state design differs from RL-AE control~\cite{lee2024learning}: the ground-truth illuminant is unavailable at deployment in RL-AWB and thus cannot appear in the state. We therefore encode rich chromatic statistics without privileged labels and add a compact history descriptor for recent adjustments.

\begin{itemize}
  \item \textbf{Illumination-related Features.}
    Following prior works~\cite{barron2015convolutional,afifi2019color,afifi2019sensor}, we represent an image with a log-chrominance (RGB-$uv$) histogram $\mathbf{H}\!\in\!\mathbb{R}^{m\times m\times 3}$ (granularity $m=60$). We then apply $\ell_1$-normalization to $\mathbf{H}$, followed by element-wise square-root, and flatten it to $\mathbf{s}_{\mathrm{WB}}\in\mathbb{R}^{3m^2}$.

  \item \textbf{Parameter-related Features.}
 To capture the trajectory of parameter adjustments, analogous to how humans consider past tuning attempts, we append a compact history vector $\mathbf{h}_s \in \mathbb{R}^{11}$ that encodes recent parameter values for both $N\%$ and $p$, along with a normalized timestep counter.

  \item \textbf{Two-branch Backbone.}
The full state is $s_t=(\mathbf{s}_{\mathrm{WB}},\mathbf{h}_s)$. Both actor and critic employ two-branch MLP encoders that map each input to 64-dimensional embeddings $\mathbf{z}_{\mathrm{WB}}$ and $\mathbf{z}_{\mathrm{hist}}$, which are then fused. The actor outputs $\boldsymbol{\mu}$ and $\log\boldsymbol{\sigma}^2 \in \mathbb{R}^2$ for reparameterized sampling of the two continuous actions, while twin critics compute Q-values with minimum selection to reduce overestimation.
\end{itemize}

\paragraph{Action Design.}
We use relative, continuous actions to jointly tune the gray-pixel percentage $N$ and Minkowski order $p$: $\text{param}_{t+1}=\text{param}_t+a_t$, with $\text{param}\in\{N,p\}$. Actions are sampled from the policy, squashed by $\tanh$ to $[-1,1]$, and rescaled to valid ranges (e.g., $a_t^{(N)}\!\in[-0.6,0.6]$, $a_t^{(p)}\!\in[-4,4]$). This yields smooth updates and coordinated adaptation.

\paragraph{Reward Design.}
We measure the quality of illuminant estimation by the angular error. To stabilize the training across images with different initial errors $E_0$, the main reward signal is the relative error improvement:
\begin{equation}
E=\arccos\!\left(\frac{\langle\hat{\mathbf e},\mathbf e\rangle}{\|\hat{\mathbf e}\|\,\|\mathbf e\|}\right),
\end{equation}
\begin{equation}
R_{\mathrm{err}}=\frac{E_0-E_t}{E_0+\left(\displaystyle\frac{E_0}{c_1}\right)^{\alpha}},
\end{equation}
where $c_1$ is the average initial error and $\alpha\!=\!0.6$. To discourage overly large moves, we add an action cost $R_{\mathrm{act}}=-\lambda\sqrt{\left(\displaystyle a_1/0.6\right)^2+\left(\displaystyle a_2/4\right)^2}$,
\noindent where $\lambda\!=\!0.1$ and $a_1,a_2$ control the gray-pixel selection percentage $N\%$ and Minkowski order $p$. A difficulty-aware relaxation scales the penalty: $R_{\text{step}}=R_{\mathrm{err}}+\left(1-\displaystyle E_0/c_2\right)\times R_{\mathrm{act}}$,
\noindent where $c_2$ is the maximum initial error. Episodes stop after three steps of estimation stability. At termination, we add bonus $R_{\rho} \in \{+50, +30, +20, +10, -10\}$ for improvement ratio $\rho = E_t/\max(E_0, 10^{-12})$ in ranges $[0, 0.8)$, $[0.8, 0.9)$, $[0.9, 0.95)$, $[0.95, 1.0)$, $\ge 1.0$, yielding $R_{\text{final}} = R_{\text{step}} + R_{\rho}$.

\paragraph{Optimization.}
We adopt the off-policy Soft Actor-Critic (SAC) algorithm~\cite{haarnoja2018sac} with a stochastic policy and critics implemented with twin Q-value heads.

\subsection{Curriculum Learning}
\label{subsec:curriculum}

\paragraph{Stage 1: Single-Image Parameter Tuning.}
Stage 1 uses a fixed training image to train the agent on error reduction through sequential parameter adjustments and termination detection. Once behavior stabilizes and the agent reliably stops at convergence, we proceed to Stage 2.

\paragraph{Stage 2: Multi-Image Tuning.}
We use a curriculum pool $\mathcal{D}_c=\{x_1,\dots,x_M\}$ ($M{=}5$). For each training data $x_i$, the agent runs 5 consecutive episodes, then cycles to $x_{i+1}$ (wrapping after $x_M$). This cyclic schedule reduces environment resets, captures short-horizon patterns in the replay buffer, and exploits SAC's off-policy experience reuse for stable updates.

\section{LEVI Dataset}
\label{sec:Multi-camera Dataset of Nighttime Color Constancy}
\begin{figure}[t]
    \centering
    \includegraphics[width=0.85\linewidth, trim={0cm 0cm 0cm 0.1cm},clip]{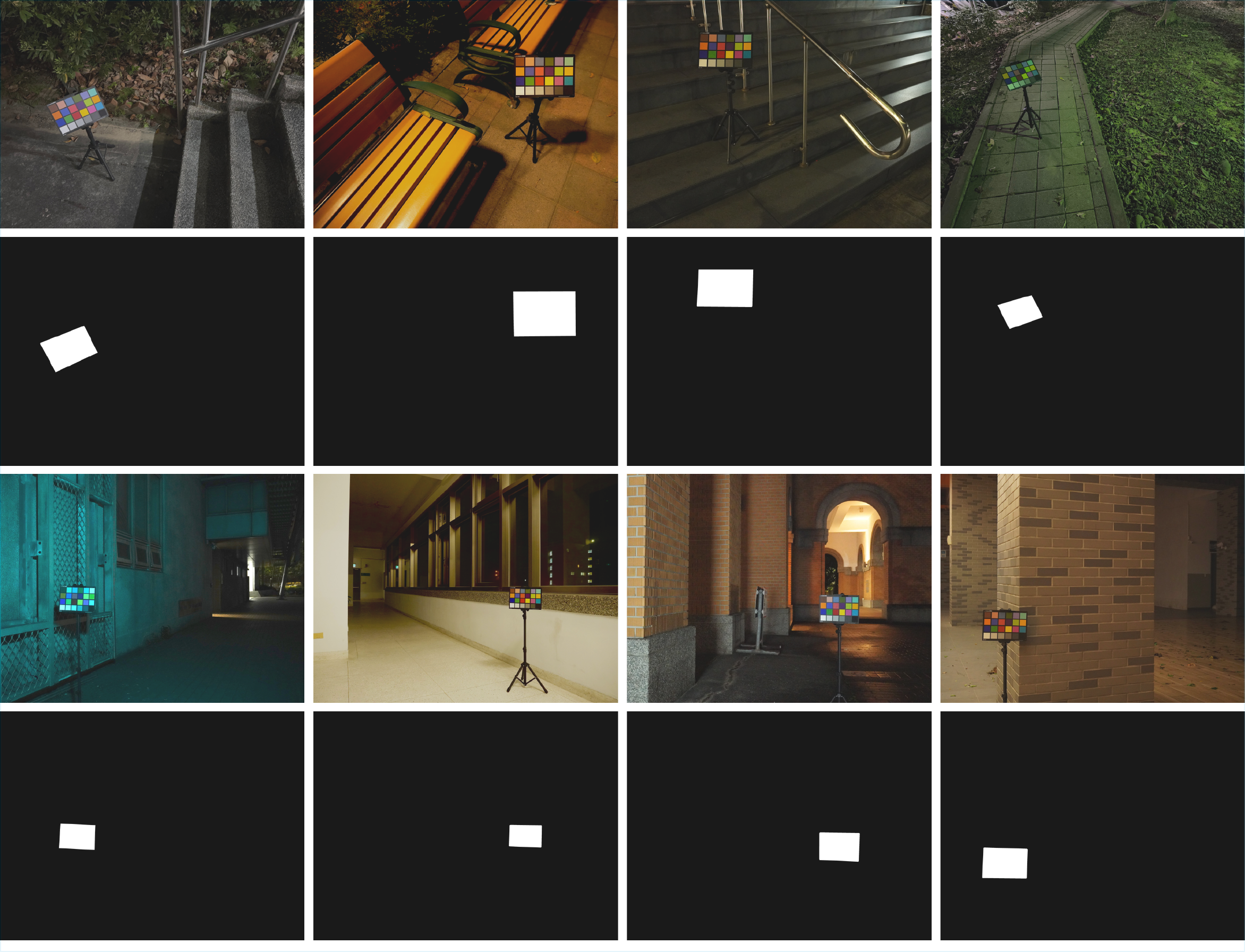}
    \caption{\textbf{Sample images from the proposed LEVI dataset with their corresponding Color Checker mask annotations.} The dataset captures diverse nighttime scenes with complex mixed lighting, low illumination, and high ISO conditions.}
    \label{fig:LEVIsample}
\end{figure}

\begin{figure}[t]
    \centering
    \begin{subfigure}[t]{0.48\linewidth}
        \centering
        \includegraphics[width=\linewidth, trim={0cm 0cm 0cm 0.60cm},clip]{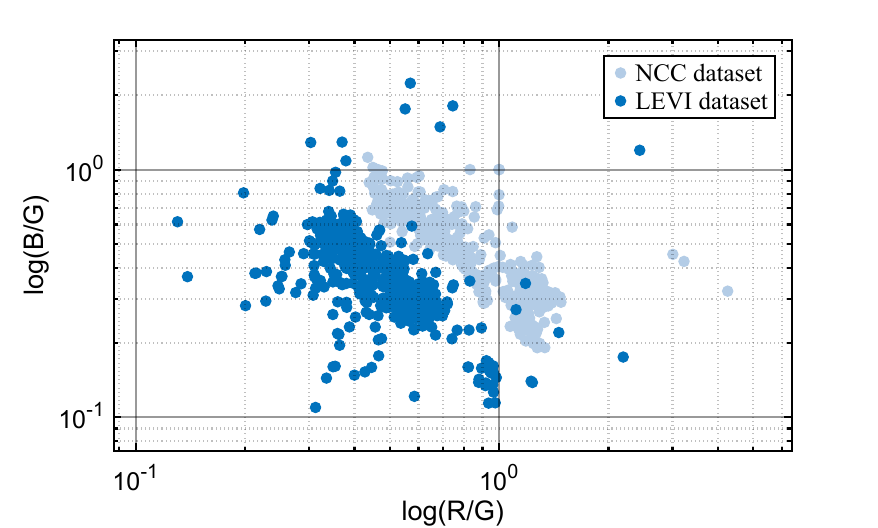}
        \caption{Illuminant distribution.}
        \label{fig:LEVIdistribution}
    \end{subfigure}
    \hfill
    \begin{subfigure}[t]{0.48\linewidth}
        \centering
        \includegraphics[width=\linewidth, trim={0cm 0.1cm 0cm 0.5cm},clip]{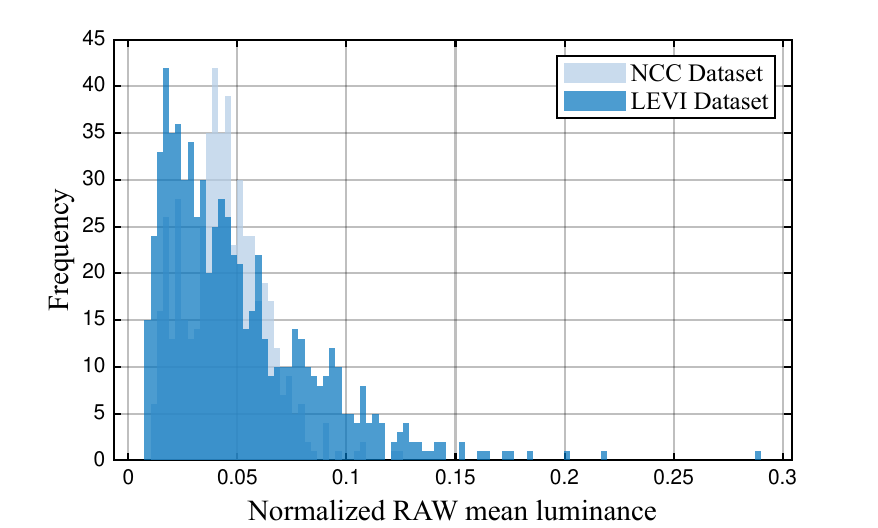}
        \caption{Normalized mean luminance histogram.}
        \label{fig:LEVIlumiDistribution}
    \end{subfigure}
    \caption{\textbf{Dataset statistics of the LEVI and NCC datasets.} (a) Illuminant distribution over all collected nighttime images. (b) Normalized mean luminance histogram showing that LEVI contains more low-luminance images.}
    \label{fig:LEVIdatasetStats}
\end{figure}

Prior to our work, NCC dataset~\cite{cheng2024nighttime} was the only public nighttime color constancy benchmark, containing 513 images from a single camera. To enable cross-sensor evaluation, we introduce the \textbf{L}ow-light \textbf{E}vening \textbf{V}ision \textbf{I}llumination (\textbf{LEVI}) dataset. The first multi-camera nighttime benchmark comprising 700 linear RAW images from two systems: iPhone 16 Pro (images \#1--370, 4320$\times$2160, 12-bit) and Sony ILCE-6400 (images \#371--700, 6000$\times$4000, 14-bit), with ISO ranging from 500 to 16,000. Each scene contains a Macbeth Color Checker with manual annotations. Ground-truth illuminants are computed as median RGB values of non-saturated achromatic patches. All images are black-level corrected and converted to linear RGB. \cref{fig:LEVIsample} shows sample images with Color Checker masks;
\cref{fig:LEVIdistribution} and \cref{fig:LEVIlumiDistribution} compare the illuminant and luminance distributions of NCC and LEVI datasets. LEVI complements NCC by covering broader lighting conditions and containing more low-luminance nighttime images, offering a new benchmark for low-light color constancy evaluation.

\section{Experiments}
\label{sec:experiments}

\subsection{Implementation Details}
\label{subsubsec:exp-impl}
We normalize image resolutions across datasets by downsampling: iPhone 16 Pro captures in LEVI are resized by 0.25$\times$, while Sony ILCE-6400 images in LEVI and all NCC images are resized by 0.125$\times$. The performance of color constancy methods is evaluated by the standard angular error metric (measured in degrees). RL-AWB (SAC) is trained on an Intel Core i5-13600K CPU. Training batch size is 256, $\gamma = 0.1$, $\tau = 0.005$, learning rate $3\times10^{-4}$, and 16 parallel environments over 15,000 timesteps, with updates starting after 100 initial steps. At inference, the agent iteratively updates parameters until convergence (average 3 iterations), with a runtime of approximately 1.1\,s per image ($\sim$360\,ms per call) on an NVIDIA RTX~3080 GPU.

\subsection{Results and Comparisons}
\label{subsec:exp-result}

Experimental results in \cref{tab:statistical_results} show that the proposed SGP-LRD consistently outperforms other statistical baselines, yielding the minimum angular errors on both NCC and LEVI datasets. Following our ablation study (\cref{subsec:exp-ablation}), we train RL-AWB using only 5 images per dataset. In contrast, all learning-based baselines (FFCC ~\cite{barron2017fast}, C$^4$~\cite{yu2020cascading}, C$^5$~\cite{afifi2021cross}, FC$^4$~\cite{hu2017fc4}, PCC~\cite{Wei2023ColorCF}, GCC~\cite{chang2025gcc}, ePCC~\cite{liu2026color}) are trained on the full training sets according to their official three-fold cross-validation protocols. As shown in \cref{tab:cross_dataset_vertical_tsi}, our proposed RL-AWB framework further enhances illuminant estimation performance. Despite the minimal supervision (5 training images), RL-AWB remains competitive with fully-trained models, showcasing a superior accuracy-data trade-off in the few-shot regime.

\begin{figure}[htbp]
    \centering
    \includegraphics[width=\linewidth, trim={0cm 0.1cm 0cm 0cm},clip]{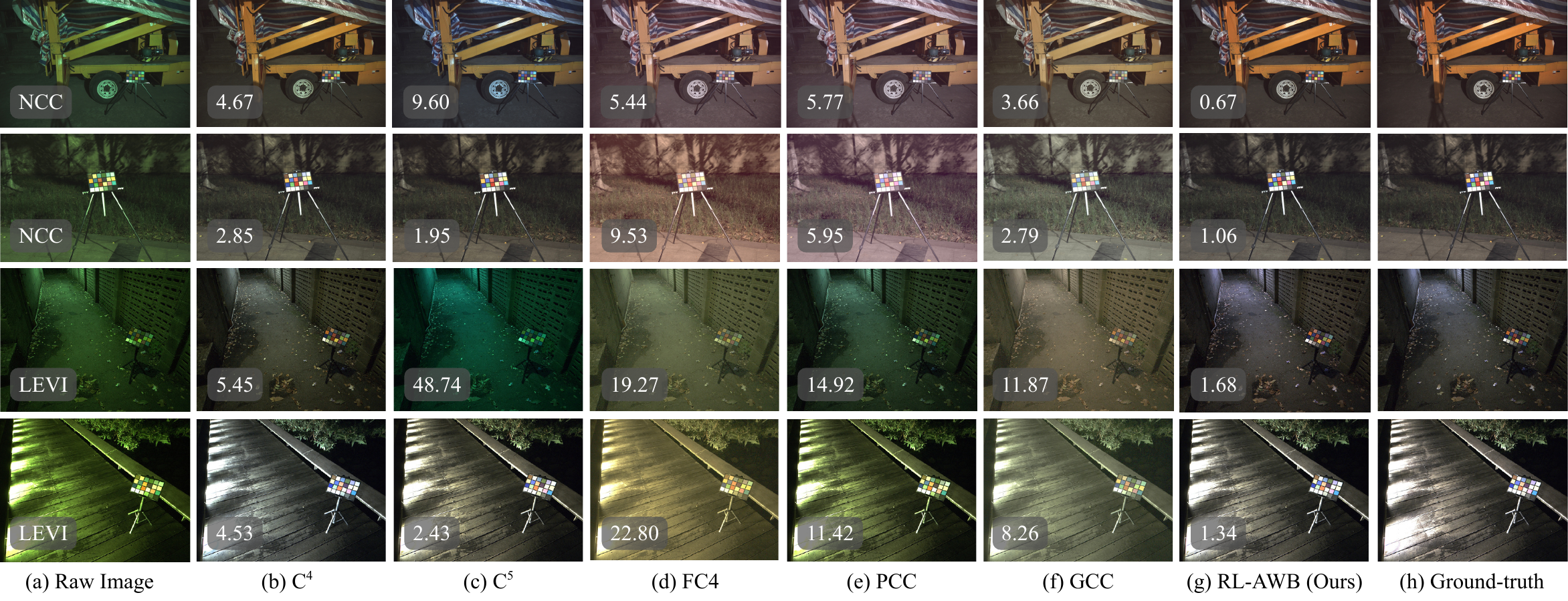}
    \caption{\textbf{Visual comparison of cross-dataset performance under domain shift.} Top: LEVI → NCC; Bottom: NCC → LEVI. Unlike prior learning-based approaches that struggle to generalize to unseen sensor domains, RL-AWB shows superior robustness across diverse sensor characteristics. Images are gamma-corrected for visualization.}
    \label{fig:comparison}
\end{figure}

Existing learning-based baselines suffer from a substantial performance degradation under cross-dataset evaluation. Specifically, for both NCC$\rightarrow$ LEVI and LEVI$\rightarrow$ NCC, the median and worst-25\% errors increase markedly compared to their in-dataset counterparts. This degradation highlights the severe impact of domain and sensor shifts, as the two datasets differ significantly in both scene content and camera characteristics, hindering the generalization of models trained on a single domain. Rather than directly regressing illuminant RGB as in deep learning-based methods, RL-AWB focuses on adaptively tuning the control parameters of SGP-LRD on a per-image basis. Illuminant estimation is then performed by the underlying statistical model, whose inherent robustness to distribution shifts supports reliable cross-dataset generalization. These results show that combining statistical estimation with reinforcement learning effectively mitigates the severe generalization degradation typical of purely learning-based methods. \cref{fig:comparison} shows qualitative results of cross-sensor performance on several nighttime scenes from NCC and LEVI datasets.

\begin{table}[t]
\centering
\fontsize{8.5pt}{10pt}\selectfont
\caption{\textbf{Evaluation of statistics-based methods on both the NCC and LEVI datasets.} Angular error in degrees.}
\label{tab:statistical_results}


\resizebox{\textwidth}{!}{%
\begin{tabular}{p{2.6cm} | p{0.95cm} p{0.95cm} p{0.95cm} p{0.95cm} p{0.95cm} | p{1.0cm} p{0.95cm} p{0.95cm} p{0.95cm} p{0.95cm}}
\toprule
\multicolumn{1}{l|}{}
& \multicolumn{5}{c|}{NCC} & \multicolumn{5}{c}{LEVI} \\
\cmidrule(lr){2-6} \cmidrule(lr){7-11}
{Method} &
\makebox[1.04cm]{Med.} & \makebox[1.04cm]{Mean} & \makebox[1.04cm]{Tri.} & \makebox[1.04cm]{B-25} & \makebox[1.04cm]{W-25} & \makebox[1.04cm]{Med.} & \makebox[1.04cm]{Mean} & \makebox[1.04cm]{Tri.} & \makebox[1.04cm]{B-25} & \makebox[1.04cm]{W-25} \\

\midrule

GE-1st ~\cite{van2007edge}& \makebox[1.05cm]{4.14} & \makebox[1.05cm]{5.17} & \makebox[1.05cm]{4.35} & \makebox[1.05cm]{1.25} & \makebox[1.05cm]{10.87} & \makebox[1.05cm]{3.94} & \makebox[1.05cm]{4.31} & \makebox[1.05cm]{3.97} & \makebox[1.05cm]{1.82} & \makebox[1.05cm]{7.45} \\
GE-2nd ~\cite{van2007edge}& \makebox[1.05cm]{3.58} & \makebox[1.05cm]{4.64} & \makebox[1.05cm]{3.78} & \makebox[1.05cm]{1.11} & \makebox[1.05cm]{9.93} & \makebox[1.05cm]{4.17} & \makebox[1.05cm]{4.49} & \makebox[1.05cm]{4.19} & \makebox[1.05cm]{1.80} & \makebox[1.05cm]{7.76} \\
MSGP ~\cite{Qian2019Revisiting} 
& \cellcolor{lightyellow}\makebox[1.05cm]{2.48} 
& \cellcolor{lightyellow}\makebox[1.05cm]{3.52} 
& \cellcolor{lightyellow}\makebox[1.05cm]{2.70} 
& \cellcolor{lightyellow}\makebox[1.05cm]{0.80} 
& \makebox[1.05cm]{8.02} 
& \cellcolor{lightyellow}\makebox[1.05cm]{3.12} 
& \cellcolor{lightorange}\makebox[1.05cm]{3.34} 
& \cellcolor{lightorange}\makebox[1.05cm]{3.14} 
& \cellcolor{lightyellow}\makebox[1.05cm]{1.54} 
& \cellcolor{lightorange}\makebox[1.05cm]{5.52} \\
GI ~\cite{qian2019finding} 
& \makebox[1.05cm]{3.13} 
& \makebox[1.05cm]{4.52} 
& \makebox[1.05cm]{3.40} 
& \makebox[1.05cm]{0.91} 
& \makebox[1.05cm]{10.60} 
& \cellcolor{lightorange}\makebox[1.05cm]{3.10} 
& \cellcolor{lightyellow}\makebox[1.05cm]{3.42} 
& \cellcolor{lightyellow}\makebox[1.05cm]{3.15} 
& \cellcolor{lightorange}\makebox[1.05cm]{1.49} 
& \cellcolor{lightyellow}\makebox[1.05cm]{5.91} \\
BCC ~\cite{Ulucan2023BlockbasedCC} & \makebox[1.05cm]{3.06} & \makebox[1.05cm]{3.81} & \makebox[1.05cm]{3.23} & \makebox[1.05cm]{1.05} & \cellcolor{lightorange}\makebox[1.05cm]{7.78} & \makebox[1.05cm]{4.23} & \makebox[1.05cm]{4.53} & \makebox[1.05cm]{4.28} & \makebox[1.05cm]{2.52} & \makebox[1.05cm]{7.06} \\
RGP ~\cite{cheng2024nighttime} 
& \cellcolor{lightorange}\makebox[1.05cm]{2.22} 
& \cellcolor{lightorange}\makebox[1.05cm]{3.33} 
& \cellcolor{lightorange}\makebox[1.05cm]{2.44} 
& \cellcolor{lightred}\makebox[1.05cm]{0.68} 
& \cellcolor{lightyellow}\makebox[1.05cm]{7.81} 
& \makebox[1.05cm]{3.21} 
& \makebox[1.05cm]{3.56} 
& \makebox[1.05cm]{3.29} 
& \makebox[1.05cm]{1.63} 
& \makebox[1.05cm]{6.12} \\

SGP-LRD (Ours) 
& \cellcolor{lightred}\makebox[1.05cm]{2.12} 
& \cellcolor{lightred}\makebox[1.05cm]{3.11} 
& \cellcolor{lightred}\makebox[1.05cm]{2.29} 
& \cellcolor{lightred}\makebox[1.05cm]{0.68} 
& \cellcolor{lightred}\makebox[1.05cm]{7.22} 
& \cellcolor{lightred}\makebox[1.05cm]{3.08} 
& \cellcolor{lightred}\makebox[1.05cm]{3.25} 
& \cellcolor{lightred}\makebox[1.05cm]{3.07} 
& \cellcolor{lightred}\makebox[1.05cm]{1.40} 
& \cellcolor{lightred}\makebox[1.05cm]{5.46} \\
\bottomrule
\end{tabular}%
}

\normalsize
\end{table}

\begin{table}[t]
\centering
\fontsize{8.5pt}{10pt}\selectfont
\caption{\textbf{Cross-dataset evaluation of the learning-based methods between NCC and LEVI datasets.} All learning-based baselines are implemented using three-fold cross-validation protocols and trained on the complete dataset.}
\label{tab:cross_dataset_vertical_tsi}


\resizebox{\textwidth}{!}{%
\begin{tabular}{p{2.6cm} | p{0.95cm} p{0.95cm} p{0.95cm} p{0.95cm} p{0.95cm} | p{0.95cm} p{0.95cm} p{0.95cm} p{0.95cm} p{0.95cm}}
\toprule
\multicolumn{1}{l|}{Train $\rightarrow$ Test}
& \multicolumn{5}{c|}{NCC $\rightarrow$ LEVI} & \multicolumn{5}{c}{LEVI $\rightarrow$ NCC} \\
\cmidrule(lr){2-6} \cmidrule(lr){7-11}
{Method} &
\makebox[1.04cm]{Med.} & \makebox[1.04cm]{Mean} & \makebox[1.04cm]{Tri.} & \makebox[1.04cm]{B-25} & \makebox[1.04cm]{W-25} & \makebox[1.04cm]{Med.} & \makebox[1.04cm]{Mean} & \makebox[1.04cm]{Tri.} & \makebox[1.04cm]{B-25} & \makebox[1.04cm]{W-25} \\

\midrule

FC$^4$ ~\cite{hu2017fc4}
& \makebox[1.05cm]{11.8} & \makebox[1.05cm]{12.3} & \makebox[1.05cm]{11.9} & \makebox[1.05cm]{6.36} & \makebox[1.05cm]{19.2}
& \makebox[1.05cm]{13.2} & \makebox[1.05cm]{14.4} & \makebox[1.05cm]{13.8} & \makebox[1.05cm]{6.17} & \makebox[1.05cm]{24.1} \\

FFCC ~\cite{barron2017fast}
& \cellcolor{lightyellow}\makebox[1.05cm]{4.44}
& \cellcolor{lightyellow}\makebox[1.05cm]{7.12}
& \cellcolor{lightyellow}\makebox[1.05cm]{5.17}
& \cellcolor{lightyellow}\makebox[1.05cm]{1.96}
& \makebox[1.05cm]{16.7}
& \makebox[1.05cm]{7.54}
& \makebox[1.05cm]{8.51}
& \makebox[1.05cm]{7.60}
& \makebox[1.05cm]{4.36}
& \makebox[1.05cm]{14.8} \\

$C^4$ ~\cite{yu2020cascading}
& \cellcolor{lightorange}\makebox[1.05cm]{3.09} & \cellcolor{lightorange}\makebox[1.05cm]{3.47} & \cellcolor{lightorange}\makebox[1.05cm]{3.18} & \cellcolor{lightred}\makebox[1.05cm]{1.18} & \cellcolor{lightorange}\makebox[1.05cm]{6.37}
& \cellcolor{lightyellow}\makebox[1.05cm]{5.85} & \cellcolor{lightyellow}\makebox[1.05cm]{6.73} & \cellcolor{lightyellow}\makebox[1.05cm]{6.04} & \makebox[1.05cm]{2.86} & \cellcolor{lightyellow}\makebox[1.05cm]{12.2} \\

$C^5$ ~\cite{afifi2021cross}
& \makebox[1.05cm]{9.12} & \makebox[1.05cm]{11.7} & \makebox[1.05cm]{9.85} & \makebox[1.05cm]{3.76} & \makebox[1.05cm]{23.4}
& \cellcolor{lightorange}\makebox[1.05cm]{4.47} & \cellcolor{lightorange}\makebox[1.05cm]{5.46} & \cellcolor{lightorange}\makebox[1.05cm]{4.68} & \cellcolor{lightorange}\makebox[1.05cm]{1.70} & \cellcolor{lightorange}\makebox[1.05cm]{10.9} \\

PCC ~\cite{Wei2023ColorCF}
& \makebox[1.05cm]{11.1} & \makebox[1.05cm]{12.5} & \makebox[1.05cm]{11.7} & \makebox[1.05cm]{7.36} & \makebox[1.05cm]{19.7}
& \makebox[1.05cm]{8.96} & \makebox[1.05cm]{10.4} & \makebox[1.05cm]{9.35} & \makebox[1.05cm]{6.24} & \makebox[1.05cm]{16.7} \\

GCC ~\cite{chang2025gcc}
& \makebox[1.05cm]{28.1} & \makebox[1.05cm]{43.0} & \makebox[1.05cm]{40.5} & \makebox[1.05cm]{12.2} & \makebox[1.05cm]{90.0}
& \makebox[1.05cm]{9.77} & \makebox[1.05cm]{41.1} & \makebox[1.05cm]{28.1} & \cellcolor{lightyellow}\makebox[1.05cm]{2.26} & \makebox[1.05cm]{90.0} \\

ePCC ~\cite{liu2026color}
& \makebox[1.05cm]{9.59} 
& \makebox[1.05cm]{10.2} 
& \makebox[1.05cm]{9.67} 
& \makebox[1.05cm]{7.05} 
& \cellcolor{lightyellow}\makebox[1.05cm]{14.3}
& \makebox[1.05cm]{7.92} 
& \makebox[1.05cm]{8.77} 
& \makebox[1.05cm]{8.04} 
& \makebox[1.05cm]{5.01} 
& \makebox[1.05cm]{13.9} \\

RL-AWB (Ours)
& \cellcolor{lightred}\makebox[1.05cm]{3.03} & \cellcolor{lightred}\makebox[1.05cm]{3.24} & \cellcolor{lightred}\makebox[1.05cm]{3.04} & \cellcolor{lightorange}\makebox[1.05cm]{1.45} & \cellcolor{lightred}\makebox[1.05cm]{5.36}
& \cellcolor{lightred}\makebox[1.05cm]{1.99} & \cellcolor{lightred}\makebox[1.05cm]{3.12} & \cellcolor{lightred}\makebox[1.05cm]{2.25} & \cellcolor{lightred}\makebox[1.05cm]{0.67} & \cellcolor{lightred}\makebox[1.05cm]{7.39} \\
\bottomrule
\end{tabular}%
}

\normalsize
\end{table}

\begin{table}[t]
\centering
\fontsize{8.5pt}{10pt}\selectfont
\caption{\textbf{Evaluation results on Gehler-Shi dataset trained on NCC dataset.} F$C^4$, $C^4$, $C^5$, PCC, GCC, ePCC, and the proposed RL-AWB are trained on the NCC dataset and evaluated on the Gehler--Shi dataset. Compared with our SGP-LRD, the proposed RL-AWB framework achieves a reduction of 5.9\% in the median angular error and 9.8\% in the best-25\% angular error, showing that RL-AWB generalizes well across low-light and well-illuminated images.}
\label{tab:gehler_results}


\begin{tabular}{
p{2.5cm} |
>{\centering\arraybackslash}p{1.04cm}
>{\centering\arraybackslash}p{1.04cm}
>{\centering\arraybackslash}p{1.04cm}
>{\centering\arraybackslash}p{1.04cm}
>{\centering\arraybackslash}p{1.04cm}
}
\toprule
Method &
\makebox[1.02cm]{Med.} &
\makebox[1.02cm]{Mean} &
\makebox[1.02cm]{Tri.} &
\makebox[1.02cm]{B-25} &
\makebox[1.0cm]{W-25} \\
\midrule



FC$^4$ ~\cite{hu2017fc4}
& 13.8
& 15.8
& 14.6
& 6.38
& 18.8 \\

$C^4$ ~\cite{yu2020cascading}
& 5.62
& 6.52
& 5.84
& 2.43
& 12.0 \\

$C^5$ ~\cite{afifi2021cross}
& \cellcolor{lightyellow} 3.34
& \cellcolor{lightyellow} 3.97
& \cellcolor{lightyellow} 3.43
& \cellcolor{lightyellow} 1.32
& \cellcolor{lightred} 7.80 \\

PCC ~\cite{Wei2023ColorCF}
& 4.97
& 8.50
& 6.07
& 1.64
& 20.9 \\

GCC ~\cite{chang2025gcc}
& 8.44
& 20.2
& 9.68
& 3.06
& 58.9 \\

ePCC ~\cite{liu2026color}
& 4.10
& 5.08
& 4.11
& 1.49
& 10.7 \\

\midrule

SGP-LRD (Ours)
& \cellcolor{lightorange} 2.38 & 
\cellcolor{lightorange} 3.64 & \cellcolor{lightorange} 2.64 & \cellcolor{lightorange} 0.51 & 
\cellcolor{lightyellow} 8.89 \\

RL-AWB (Ours)
& \cellcolor{lightred} 2.24
& \cellcolor{lightred} 3.50
& \cellcolor{lightred} 2.51
& \cellcolor{lightred} 0.46
& \cellcolor{lightorange} 8.67 \\
\bottomrule
\end{tabular}

\end{table}

Beyond nighttime color constancy, we further examine whether the proposed method generalizes to daytime scenarios. To adapt SGP-LRD to well-lit datasets, we remove the local variance and the color deviation filtering modules, and subsequently perform our RL-based parameter tuning. The evaluation results on Gehler-Shi dataset~\cite{gehler2008bayesian, shi2010reprocessed} are shown in \cref{tab:gehler_results}. Despite being tailored for low-light nighttime scenes, RL-AWB achieves state-of-the-art generalization capability compared with other baselines.

\subsection{Ablation Studies}
\label{subsec:exp-ablation}

\begin{figure}[t]
    \centering
    \includegraphics[width=\linewidth]{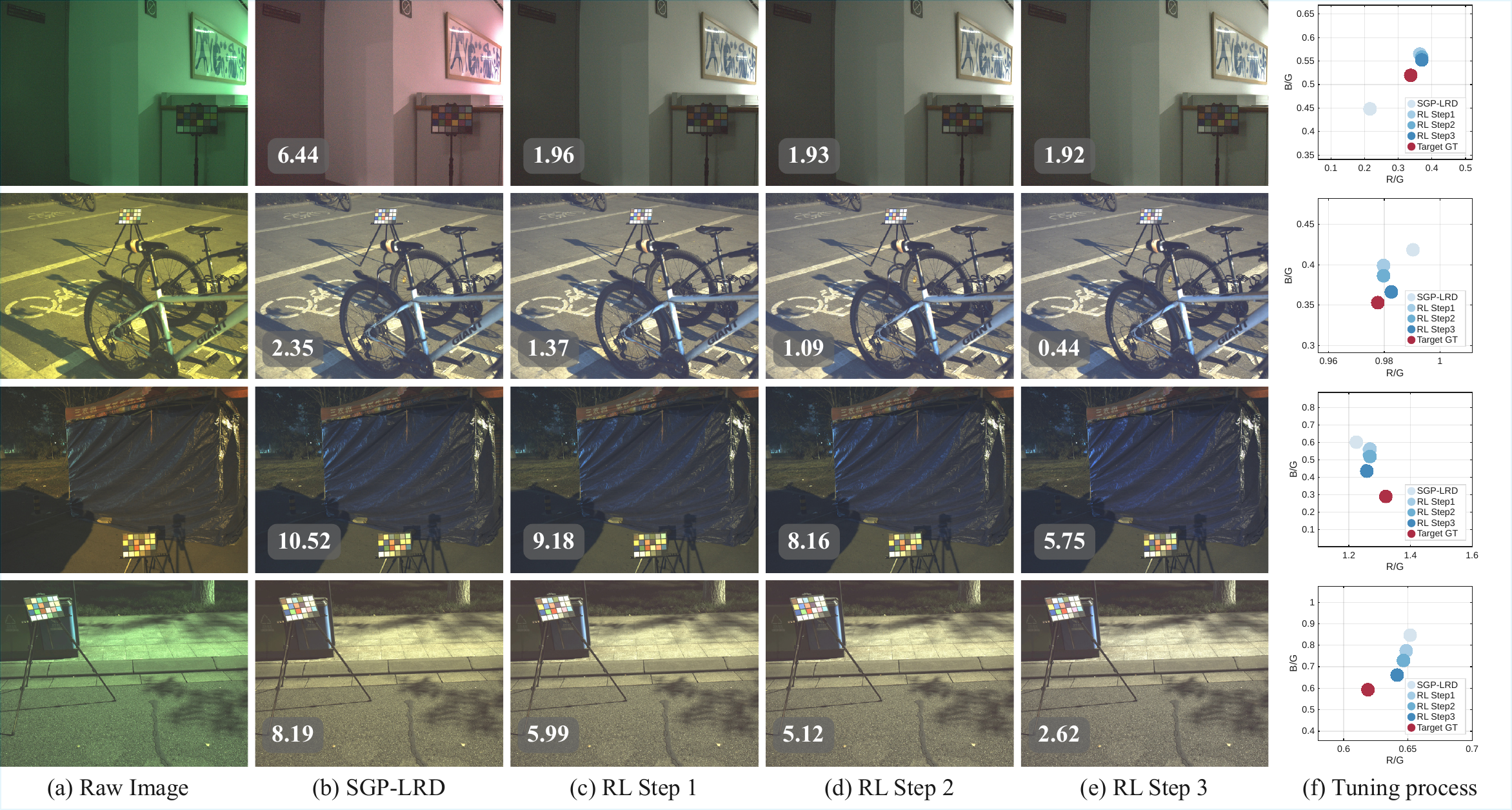}
    \caption{\textbf{Illustration of the RL-AWB auto-tuning process for representative nighttime scenes.}
    For each image, we visualize the initial input, several intermediate correction results along the trajectory of the RL policy, and the final output, together with the corresponding angular error at each step.
    As the agent iteratively updates the SGP-LRD parameters, the corrected images gradually approach the ground-truth white-balanced results, and the angular error decreases. Note that the images shown are gamma-corrected for visualization.}
    \label{fig:rl_adjust}
\end{figure}

\paragraph{Necessity of the RL framework.}
We validate the necessity of the proposed RL-AWB framework by comparing it against one-step parameter regressors. In principle, a dense per-image grid search over the algorithm parameters can provide parameters ground truth. Nonetheless, such a procedure requires illuminant ground truth at test time to select the optimal configuration via an argmin operation, making it incompatible with real-world deployment. For example, evaluating 440 parameter configurations per image (11 values for $N$ and 40 values for $p$) requires approximately 94 hours of computation. We utilize this grid-search procedure only offline to generate supervisory targets for one-step regressors. Specifically, we train both MLP and CNN models to predict $(N,p)$ directly from the same RGB-uv histogram used by RL-AWB, where the MLP backbone is identical to the SAC actor network. The regressors are trained on NCC (N), LEVI (L), and Gehler-Shi (G) datasets following the same evaluation protocol as RL-AWB, including three-fold in-domain evaluation and all cross-domain transfer settings. The median angular error results are shown in Tab.~\ref{tab:ablation_mlpcnn}. This performance advantage arises from RL-AWB’s iterative refinement mechanism. Unlike one-step regressors, which must commit to a single parameter estimate and cannot revise it afterward, the RL agent progressively adjusts $(N,p)$ over multiple steps and can recover from suboptimal intermediate decisions. Such error-correction behavior is fundamentally unavailable to one-step regression models by construction. \cref{fig:rl_adjust} shows the stepwise corrections.

\begin{table}[t]
\centering
\fontsize{8.5pt}{10pt}\selectfont
\caption{\textbf{Evaluation results on RL-AWB compared to one-step regressors.} RL-AWB consistently achieves lower final angular error than both one-step MLP and CNN regressors, despite being trained with only five images.} 
\label{tab:ablation_mlpcnn}


\begin{tabular}{l|ccc|ccc|ccc}
\toprule
& \multicolumn{3}{c|}{Train: NCC} & \multicolumn{3}{c|}{Train: LEVI} & \multicolumn{3}{c}{Train: Gehler} \\
\makebox[1.8cm][l]{Method} & 
\makebox[1.0cm]{N$\to$N} & 
\makebox[1.0cm]{N$\to$L} & 
\makebox[1.0cm]{N$\to$G} & 
\makebox[1.0cm]{L$\to$N} & 
\makebox[1.0cm]{L$\to$L} & 
\makebox[1.0cm]{L$\to$G} & 
\makebox[1.0cm]{G$\to$N} & 
\makebox[1.0cm]{G$\to$L} & 
\makebox[1.0cm]{G$\to$G} \\
\midrule
MLP & 2.01 & 3.07 & 2.29 & 2.25 & 3.02 & 2.39 & 2.20 & 3.13 & 2.34 \\
CNN & 2.06 & 3.05 & 2.34 & 2.17 & 3.04 & 2.35 & 2.07 & 3.17 & 2.30 \\
\rowcolor{black!10}
\textbf{RL-AWB} & \textbf{1.98} & \textbf{3.03} & \textbf{2.24} & \textbf{1.99} & \textbf{3.01} & \textbf{2.27} & \textbf{1.99} & \textbf{3.12} & \textbf{2.16} \\
\bottomrule
\end{tabular}
\end{table}

\begin{table}[t]
\centering
\fontsize{8.5pt}{10pt}\selectfont

\caption{\textbf{Ablation on curriculum pool size (SAC).}
The results exhibit a U-shaped performance trend. Small pools ($<5$ images) lack sufficient diversity to learn robust policies, while large pools ( $>5$ images) reduce per-sample visitation under a fixed replay budget, leading to excessive exploration noise.}
\label{tab:ablation_trainimgs}


\begin{tabular}{p{1cm} | >{\centering\arraybackslash}p{1cm} >{\centering\arraybackslash}p{1cm} >{\centering\arraybackslash}p{1cm} >{\centering\arraybackslash}p{1cm} >{\centering\arraybackslash}p{1cm} >{\centering\arraybackslash}p{1cm}}
\toprule
\multirow{2}{*}{} & \multicolumn{3}{c}{NCC} & \multicolumn{3}{c}{LEVI} \\
\cmidrule(lr){2-4} \cmidrule(lr){5-7}
 & Med. & Mean & W-25 & Med. & Mean & W-25 \\
\midrule
3  & 2.16 & 3.29 & 7.69 & 3.05 & 3.28 & 5.55 \\
\rowcolor{black!10}
\textbf{5}  & \textbf{1.98} & \textbf{3.07} & \textbf{7.22} & \textbf{3.01} & \textbf{3.22} & \textbf{5.32} \\
7  & 2.09 & 3.19 & 7.54 & 3.04 & 3.28 & 5.53 \\
9  & 2.13 & 3.23 & 7.63 & 3.03 & 3.23 & 5.39 \\
15 & 2.24 & 3.21 & 7.47 & 3.06 & 3.24 & 5.41 \\
\bottomrule
\end{tabular}
\end{table}

\paragraph{Training data number.}
We vary the Stage-2 curriculum pool size \(M\in\{3,5,7,9,15\}\) (\cref{tab:ablation_trainimgs}). We adopt \(M=5\) as the best trade-off.

\begin{table}[t]
\centering
\fontsize{8.5pt}{10pt}\selectfont

\caption{\textbf{Ablation study on SGP-LRD components.} Angular error in degrees.}
\label{tab:ablation_sgplrd}


\begin{tabular}{p{2.8cm} | >{\centering\arraybackslash}p{1cm} >{\centering\arraybackslash}p{1cm} >{\centering\arraybackslash}p{1cm} >{\centering\arraybackslash}p{1cm} >{\centering\arraybackslash}p{1cm} >{\centering\arraybackslash}p{1cm}}
\toprule
\multirow{2}{*}{} & \multicolumn{3}{c}{NCC} & \multicolumn{3}{c}{LEVI} \\
\cmidrule(lr){2-4} \cmidrule(lr){5-7}
 & \makebox[0.9cm]{Med.} & \makebox[0.9cm]{Mean} & \makebox[0.9cm]{W-25} & \makebox[0.9cm]{Med.} & \makebox[0.9cm]{Mean} & \makebox[0.9cm]{W-25} \\
\midrule
w/o Noise Filtering & \makebox[0.9cm]{2.12} & \makebox[0.9cm]{3.12} & \makebox[0.9cm]{7.32} & \makebox[0.9cm]{3.09} & \makebox[0.9cm]{3.26} & \makebox[0.9cm]{5.48} \\
w/o Color Filtering & \makebox[0.9cm]{2.51} & \makebox[0.9cm]{3.90} & \makebox[0.9cm]{9.54} & \makebox[0.9cm]{3.25} & \makebox[0.9cm]{3.68} & \makebox[0.9cm]{6.61} \\
\rowcolor{black!10}
\textbf{Full SGP-LRD} & \makebox[0.9cm]{\textbf{2.12}} & \makebox[0.9cm]{\textbf{3.11}} & \makebox[0.9cm]{\textbf{7.22}} & \makebox[0.9cm]{\textbf{3.08}} & \makebox[0.9cm]{\textbf{3.25}} & \makebox[0.9cm]{\textbf{5.46}} \\
\bottomrule
\end{tabular}
\end{table}
\paragraph{SGP-LRD.}
We ablate the two key filtering stages of SGP-LRD to assess their individual contributions. As shown in \cref{tab:ablation_sgplrd}, optimal performance is attained only when both the noise and color filtering modules are incorporated. It is worth noting that the noise filter is designed as a robustness safeguard to mitigate the impact of shot noise inherent in nighttime imagery, not an accuracy driver.



\section{Conclusion}
\label{sec:conclusion}

This study is the first to apply reinforcement learning to color constancy, showing that DRL can be effectively used for white balance tuning. Our work makes three contributions: (1) SGP-LRD, a novel statistical algorithm for nighttime color constancy, (2) RL-AWB, a reinforcement learning framework for adaptive parameter optimization, and (3) LEVI, the first multi-camera nighttime color constancy dataset. Experiments on both nighttime and daytime datasets demonstrate that the proposed method achieves competitive performance compared to existing baselines, while exhibiting strong cross-sensor robustness. Compared to one-step regressors, RL-AWB provides a practical and deployment-compatible solution that remains ground-truth free at inference time while generalizing effectively to unseen datasets and camera sensors.

\begin{figure}[t]
    \centering
    \includegraphics[width=\linewidth, trim={0cm 0.3cm 0cm 0.2cm},clip]{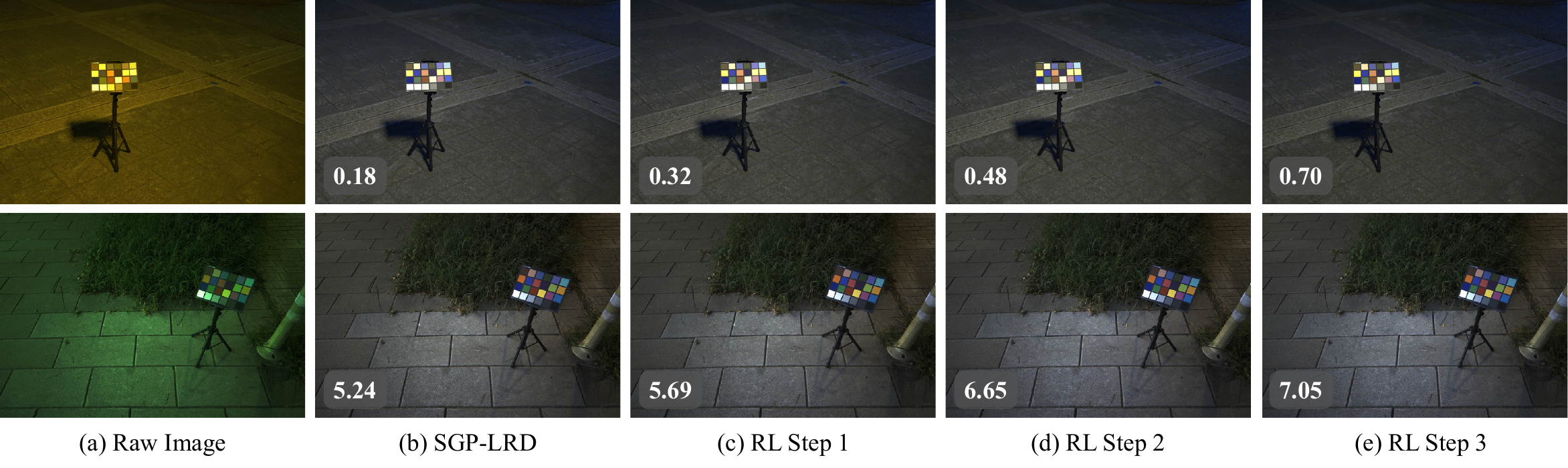}
    \caption{\textbf{Failure case analysis.} Examples of failure cases in challenging nighttime scenes, where RL-AWB over-corrects the illumination and produces visually degraded outputs. Images are gamma-corrected for visualization.}
    \label{fig:Rebuttal_Failture_Case}
\end{figure}

\paragraph{Limitations and future work.}
While RL-AWB consistently reduces overall angular error, we observe one failure mode: for images with already low initial estimation error, the RL agent may over-correct, leading to slightly degraded outputs. \cref{fig:Rebuttal_Failture_Case} illustrates such cases where the agent's parameter adjustments increase the angular error from an initial estimate. Future work will explore safety-aware reward formulations and constrained optimization strategies to mitigate over-correction, as well as hierarchical policies to efficiently coordinate additional tunable ISP parameters beyond the current two-parameter action space.

\section*{Acknowledgements}
This research was funded by the National Science and Technology Council, Taiwan, under Grants NSTC 112-2222-E-A49-004-MY2 and 113-2628-E-A49-023-. The authors are grateful to Google, NVIDIA, and MediaTek Inc. for their generous donations. Yu-Lun Liu acknowledges the Yushan Young Fellow Program by the MOE in Taiwan.

\FloatBarrier
\bibliographystyle{splncs04}
\bibliography{11_references}

\clearpage

\setcounter{section}{0}
\renewcommand{\thesection}{\Alph{section}}
\setcounter{table}{0}
\renewcommand{\thetable}{S\arabic{table}}
\setcounter{figure}{0}
\renewcommand{\thefigure}{S\arabic{figure}}
\setcounter{equation}{0}
\renewcommand{\theequation}{S\arabic{equation}}

\begin{center}
    {\Large\bfseries RL-AWB: Deep Reinforcement Learning for Auto White Balance Correction in Low-Light Nighttime Scenes\\[6pt]
    Supplementary Material\par}
\end{center}
\vspace{1em}

\section{Overview}
\label{sec:overview}
This document provides supplementary material for the paper \emph{``RL-AWB: Deep Reinforcement Learning for Auto White Balance Correction in Low-Light Nighttime Scenes''} to complement the main manuscript. \cref{sec:SGPLRDdetails} and~\cref{sec:RLAWBdetails} provide comprehensive implementation details of the proposed color constancy method, SGP-LRD, and the RL-AWB framework, respectively. \cref{sec:datasets} offers a detailed description of the proposed nighttime color constancy dataset, LEVI. Additional quantitative results are reported in \cref{sec:supresults}, while further ablation studies are presented in \cref{sec:supablation}. \cref{sec:supvisual} provides additional visual comparisons. Finally, \cref{sec:futurework} discusses future research directions.

\section{Details of SGP-LRD}
\label{sec:SGPLRDdetails}
The Laplacian of Gaussian (LoG) operator used for local contrast computation employs $\sigma=0.5$ with a dynamically computed kernel size ($\approx$7$\times$7). The key thresholds for the two-layer filtering are: variance threshold $0.025$ (LEVI) / $0.045$ (NCC); color deviation threshold $0.3$ (NCC) / $0.35$ (LEVI); and confidence threshold $= 0.9$. These values are determined via grid search on a small validation set. Following the initial gray pixel selection, we employ a two-layer filtering process to refine the candidate set, aiming to mitigate the adverse effects of sensor noise and chromatic outliers, which are prevalent in low-light imagery:

\paragraph{Noise Mitigation via Local Variance Filtering.} The first stage addresses the issue of pure noise, which produces achromatic-like responses in extremely dark areas. For each pixel (x,y) in the initial detected gray pixel mask, we compute the variance across the logarithmic RGB channels. Pixels where the intra-pixel variance is too small often lack reliable color signal and are primarily sensor noise. We then filter out these unreliable candidates:

\begin{equation}
\text{Mask}_1(x, y) = \begin{cases}
    1, & \text{if } \text{Var}\{\log(I(x, y))\} > {VarTh} \\
    0, & \text{otherwise}
\end{cases}
\end{equation}

\paragraph{Chromatic Outlier Elimination via Color Difference Filtering.} This stage aims to filter out initial detected gray pixels that are too distant from the dominant color cast of the scene's illumination. We first compute the mean logarithmic intensity for each channel of the image: $\mathbf{M} = [\bar{M}_R, \bar{M}_G, \bar{M}_B]^{\mathrm{T}}$, and then calculate the maximum absolute color deviation, $X$, for each pixel from this mean:

\begin{equation}
X(x, y) = \max_{i \in \{R, G, B\}} \left( |\log(I_{i}(x, y)) - \bar{M}_i| \right)
\end{equation}

To discard pixels that are outliers relative to the dominant scene color, we define an adaptive threshold $T_C=ColorTh \cdot \min(\textbf{M})$. Any initial detected gray pixel exceeding this deviation is removed:

\begin{equation}
\label{eq:mask2}
\text{Mask}_2(x, y) = \begin{cases}
    1, & \text{if } X(x, y) \le T_C \\
    0, & \text{otherwise}
\end{cases}
\end{equation}

Applying these two refinement layers to the initially detected gray pixels yields the Salient Gray Pixels (SGP).

\paragraph{Gray-pixel Confidence Weighting.}
A fundamental challenge in nighttime color constancy is that pixels in underexposed areas suffer from low SNR, compromising the reliability of their color measurements. To address this spatially-varying reliability, we introduce a luminance-adaptive confidence measure that weights gray-pixel candidates according to their local signal quality. For a given image with bit depth representation, we first normalize the pixel values to the range [0,1], and then the luminance map for SGPs is computed as:

\begin{equation}
LM(x) = \frac{R(x) + G(x) + B(x)}{3}
\end{equation}

\vspace{5pt}
We analyze the intensity distribution characteristics through skewness calculation on non-zero pixels. The skewness value $s_{LM}$ guides the selection of an adaptive exponent parameter E:

\begin{equation}
E = \begin{cases}
1.0 & \text{if } s_{LM} > 1.5 \\
2.0 & \text{if } 0.2 < s_{LM} \leq 1.5 \\
4.0 & \text{if } s_{LM} \leq 0.2
\end{cases}
\end{equation}

This adaptive selection responds to different scene brightness distributions: higher skewness indicates more low-intensity pixels, while lower skewness suggests more uniform illumination. The confidence weight is then computed as:

\begin{equation}
W_{SGP}(x) = 1 - \exp\left[-\left(\frac{LM(x)}{\overline{LM}}\right)^E \ \right]
\end{equation}

\paragraph{Pixel-wise Local Reflectance Difference.} We design a pixel-wise sliding window approach for local normalization. For each pixel position $(i,j)$, we define a local window $\mathcal{N}_{i,j}$ of size $w \times w$ centered at that pixel. Let $f_c(x)$ denote the detected SGPs intensity for channel $c$. We denote $\mathcal{W}_{i,j}^*$ as the set of non-zero pixels within the local window $\mathcal{W}_{i,j}$. The pixel-wise normalized local reflectance difference at position $(i,j)$ is defined as:

\begin{equation}
N_c^{i,j} = \begin{cases}
\dfrac{\left[\sum\limits_{x \in \mathcal{W}_{i,j}^*} f_c(x)\right] / \ |\mathcal{W}_{i,j}^*|}{\max\limits_{x \in \mathcal{W}_{i,j}} f_c(x)} & \text{if } \max\limits_{x \in \mathcal{W}_{i,j}} f_c(x) > 0 \\[0.8em]
\ 0 & \text{otherwise}
\end{cases}
\end{equation}

\noindent where the numerator represents the mean of non-zero elements in the local window, computed as the sum of non-zero pixels divided by their count $|\mathcal{W}_{i,j}^*|$, and the denominator is the maximum value within the window. This formulation ensures that each pixel's local context is normalized by its surrounding maximum, providing a spatially adaptive measure of local reflectance variation.

Integrating the gray-pixel confidence weights into the estimation framework, we formulate the illuminant estimation as:

\vspace{-3pt}

\begin{equation}
\hat{e}_c = \left[\frac{\sum_{(i,j) \in \Omega} \left(\mu_c^{i,j} \cdot W_{SGP}(i,j)\right)^p}{\sum_{(i,j) \in \Omega} \left(N_c^{i,j} \cdot W_{SGP}(i,j)\right)^p}\right]^{1/p}
\end{equation}

\vspace{2pt} 

\noindent where $\Omega$ represents all valid pixel positions (where $N_c^{i,j} > 0$), $W(i,j)$ is the gray-pixel confidence weight at position $(i,j)$, and $p$ is the Minkowski norm parameter. The gray-pixel confidence weight $W(i,j)$ ensures that pixels in well-illuminated, reliable regions contribute more to the final estimate, while uncertain or poorly-lit regions have reduced influence. Finally, the estimated illuminant vector is then normalized:

\vspace{-2pt}

\begin{equation}
\mathbf{\hat{e}} = \frac{(\hat{e}_R, \hat{e}_G, \hat{e}_B)}{\|(\hat{e}_R, \hat{e}_G, \hat{e}_B)\|}
\end{equation}

\vspace{2pt}

The pixel-wise sliding window strategy is designed to exploit the spatial distribution characteristics of reliable SGPs in natural scenes.

\section{Details of RL-AWB}
\label{sec:RLAWBdetails}
\paragraph{Python Environment Setting.}
We implement RL-AWB in Python, using Stable-Baselines3 to construct and train the SAC agent, and PyTorch to define custom policy and value networks. The SAC networks are implemented as dual-head multi-layer perceptrons (MLPs)
built mainly from fully connected layers. During SAC training, we rely solely on the CPU. Our experiments are run on a machine equipped with an Intel Core i5-13600K processor and 40\,GB of system memory. During environment interaction, the input images are processed by the SGP-LRD algorithm, for which we leverage a single NVIDIA RTX~3080 GPU with 10\,GB of VRAM to accelerate the computations.

\paragraph{Soft Actor-Critic (SAC) Algorithm Optimization.}
We adopt Soft Actor-Critic (SAC) to optimize the parameters of our AWB algorithm. SAC is an off-policy deep reinforcement learning method whose core idea is to jointly maximize the expected return and the policy entropy, thereby encouraging exploration and improving training stability. It follows an actor--critic architecture with two critic networks \(Q_{\theta_1}, Q_{\theta_2}\) and a stochastic actor \(\pi_\phi\). The critic loss is defined as
\begin{equation}
\begin{aligned}
J_Q(\theta_i)
 &= \mathbb{E}_{(s_t, a_t, r_t, s_{t+1}) \sim D}
    \Bigl[
      \tfrac{1}{2}
      \bigl(Q_{\theta_i}(s_t, a_t) - y_t\bigr)^2
    \Bigr], \ i \in \{1, 2\}.
\end{aligned}
\end{equation}
where \(D\) denotes the replay buffer. The target value \(y_t\) is given by
\begin{equation}
\begin{aligned}
y_t
 &= r_t + \gamma\,
    \mathbb{E}_{a_{t+1} \sim \pi_\phi} \Bigl[
      \min_{j \in \{1,2\}} Q_{\theta_j}(s_{t+1}, a_{t+1}) - \alpha \log \pi_\phi(a_{t+1} \mid s_{t+1})
    \Bigr].
\end{aligned}
\end{equation}

Here, \(\gamma \in (0,1)\) is the discount factor, \(\alpha\) is the temperature parameter that controls the strength of exploration, and \(r_t, s_t, a_t\) are the reward, state, and action at time step \(t\), respectively. Using \(\min(Q_{\theta_1}, Q_{\theta_2})\) in the target mitigates overestimation of the Q-value and stabilizes learning. The entropy term \(-\alpha \log \pi_\phi(a_{t+1} \mid s_{t+1})\) encourages the policy to remain sufficiently stochastic. The actor is updated by minimizing
\begin{equation}
  J_\pi(\phi)
  = \mathbb{E}_{s_t \sim D,\, \varepsilon \sim \mathcal{N}}
    \bigl[
      \alpha \log \pi_\phi(a_t \mid s_t)
      - Q_{\bar{\theta}}(s_t, a_t)
    \bigr]
\end{equation}
where \(a_t\) is sampled via the reparameterization trick from a noise variable \(\varepsilon\), and \(Q_{\bar{\theta}}\) denotes a slowly updated target critic. This objective strikes a balance between achieving high Q-values and preserving sufficient randomness for exploration: if a certain action yields a large Q-value, the actor increases its probability; however, when the policy becomes overly deterministic, the entropy term acts as a regularizer that penalizes low-entropy behavior and enforces diversity in the actions.

\section{Datasets}
\label{sec:datasets}
\paragraph{Low-light Evening Vision Illumination (LEVI) Dataset.}

\begin{figure}[t]
    \centering
    \includegraphics[width=\linewidth]{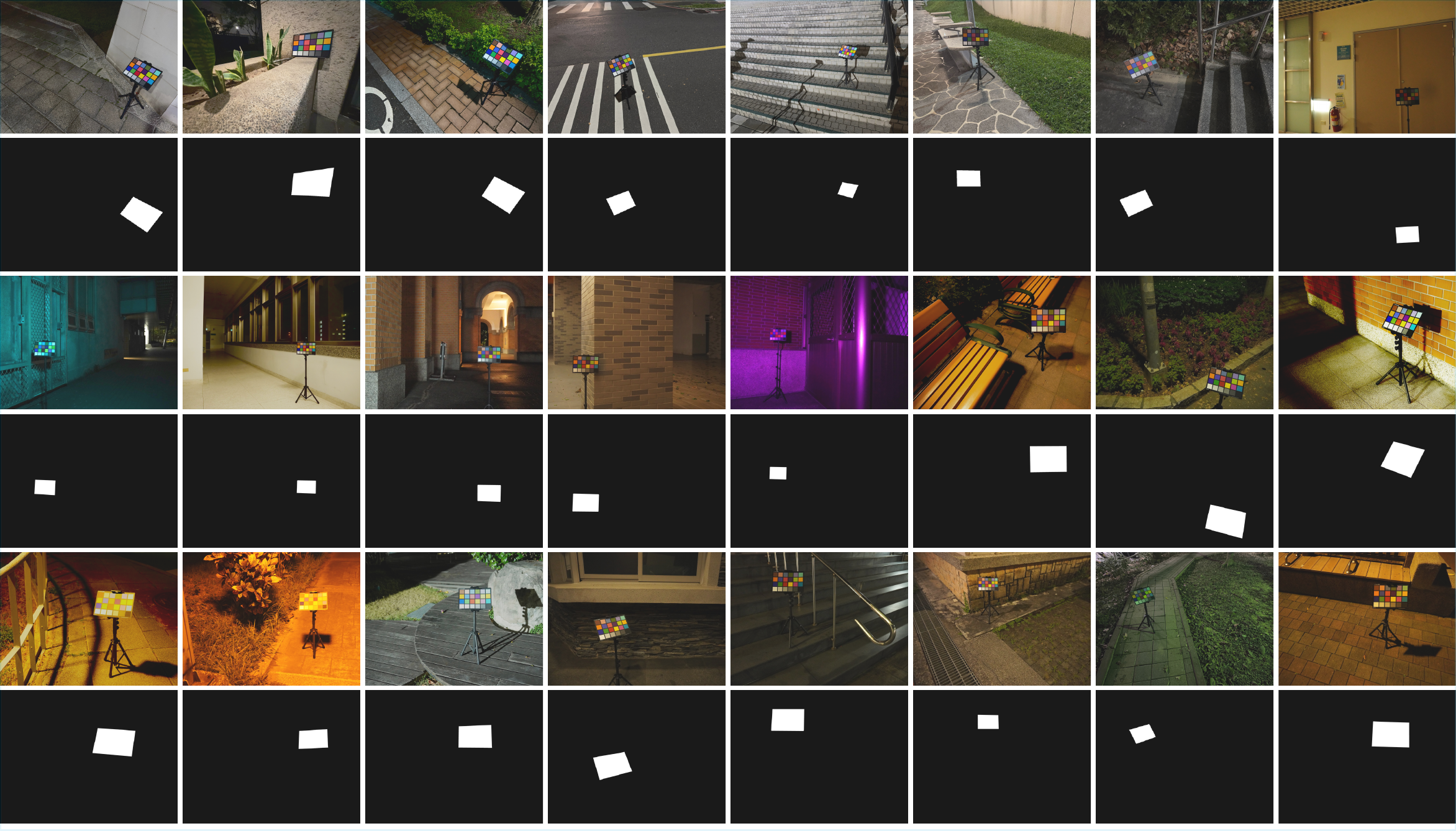}
    \caption{\textbf{Example nighttime scenes from the LEVI dataset.} LEVI covers diverse nighttime environments and illuminant conditions across multiple camera sensors.}
    \label{fig:LEVIfull}
\end{figure}

Given the limited availability of nighttime color constancy datasets, we constructed a new dataset to facilitate research in this challenging domain. Some samples are shown in \cref{fig:LEVIfull}. Prior to our work, the NCC dataset was the only publicly available dataset specifically designed for nighttime illuminant estimation, containing 513 nighttime images with corresponding ground-truth illuminants. While the NCC dataset has been valuable for initial algorithmic development, it was captured using a single camera model, limiting its utility for evaluating cross-sensor generalization, a critical requirement for practical AWB systems that must operate across diverse imaging devices.
Our dataset addresses this limitation and introduces several key improvements. First, it is the first nighttime color constancy dataset captured with multiple camera systems, enabling rigorous evaluation of cross-sensor generalization performance. The dataset comprises 700 nighttime images: images \#1--\#370 were captured using an iPhone 16 Pro at a resolution of 4320 $\times$ 2160 pixels with 12-bit depth, while images \#371--\#700 were captured using a Sony ILCE-6400 at 6000 $\times$ 4000 pixels with 14-bit depth. The ISO values range from approximately 500 to 16,000, covering a wide spectrum of low-light conditions commonly encountered in nighttime photography. Second, we provide comprehensive metadata for each image, including focal length (mm), F-number, exposure time (s), and ISO settings. This additional information enables researchers to analyze the relationship between camera settings and illuminant estimation performance, potentially leading to more robust AWB algorithms that can adapt to different capture conditions.

\section{Additional Quantitative Results}
\label{sec:supresults}
\subsection{Reproduction Angular Error}
In addition to the recovery angular error used in the main paper, we also adopt the \emph{reproduction angular error} to evaluate the discrepancy between the predicted illuminant $\hat{\mathbf e}$ and the ground-truth illuminant $\mathbf e$. We first compute the per-channel ratio vector
\begin{equation}
    LL = \frac{\mathbf e}{\hat{\mathbf e}},
\end{equation}
and then measure the angle between $LL$ and the ideal white vector
$[1,1,1]^\top$:
\begin{equation}
    \mathrm{rep}
    = \arccos\!\left(
        \frac{\langle LL,\,[1,1,1] \rangle}
             {\sqrt{3}\,\|LL\|_2}
    \right).
\end{equation}
Smaller angles indicate that the corrected image is closer to perceptual white. These two metrics complement each other: recovery angular error reflects the accuracy of illuminant estimation, while reproduction angular error reflects the perceptual quality after AWB correction. In the main paper, we report the primary comparisons using the recovery angular error; in this supplementary material, we additionally provide results under the reproduction angular error to complete the performance evaluation.

\subsection{In-dataset Quantitative Comparison}

\cref{tab:training_results,tab:sup_training_results} present the in-dataset recovery and reproduction angular errors on the NCC and LEVI datasets, respectively. All learning-based methods are trained using their official 3-fold cross-validation protocols with full training data, while RL-AWB uses only 5 training images per dataset. 
\begin{table}[t]
\centering
\fontsize{8.5pt}{10pt}\selectfont
\caption{\textbf{In-dataset evaluation results on NCC and LEVI datasets.} Recovery angular error in degrees. All learning-based baselines are implemented using three-fold cross-validation protocols and trained on the complete dataset; RL-AWB uses only 5 training images.}
\label{tab:training_results}

\vspace{-5pt}

\resizebox{\textwidth}{!}{%
\begin{tabular}{p{2.6cm} | >{\centering\arraybackslash}p{0.95cm} >{\centering\arraybackslash}p{0.95cm} >{\centering\arraybackslash}p{0.95cm} >{\centering\arraybackslash}p{0.95cm} >{\centering\arraybackslash}p{0.95cm} | >{\centering\arraybackslash}p{0.95cm} >{\centering\arraybackslash}p{0.95cm} >{\centering\arraybackslash}p{0.95cm} >{\centering\arraybackslash}p{0.95cm} >{\centering\arraybackslash}p{0.95cm}}
\toprule
\multicolumn{1}{l|}{}
& \multicolumn{5}{c|}{NCC Dataset} & \multicolumn{5}{c}{LEVI Dataset} \\
\cmidrule(lr){2-6} \cmidrule(lr){7-11}
{Method} & Med. & Mean & Tri. & B-25 & W-25 & Med. & Mean & Tri. & B-25 & W-25 \\

\midrule

FFCC ~\cite{barron2017fast}
& \makebox[1.05cm]{2.40}
& \makebox[1.05cm]{3.81}
& \makebox[1.05cm]{2.62}
& \makebox[1.05cm]{0.93}
& \makebox[1.05cm]{9.16}
& \makebox[1.05cm]{2.33}
& \makebox[1.05cm]{4.67}
& \makebox[1.05cm]{2.66}
& \makebox[1.05cm]{0.69}
& \makebox[1.05cm]{12.95}
\\

FC$^4$ ~\cite{hu2017fc4} & \makebox[1.05cm]{12.0} & \makebox[1.05cm]{11.8} & \makebox[1.05cm]{11.8} & \makebox[1.05cm]{4.70} & \makebox[1.05cm]{19.0} & \makebox[1.05cm]{7.24} & \makebox[1.05cm]{8.25} & \makebox[1.05cm]{7.26} & \makebox[1.05cm]{3.20} & \makebox[1.05cm]{15.1}
\\

C$^4$ ~\cite{yu2020cascading}
& \cellcolor{lightorange}\makebox[1.05cm]{1.66}
& \cellcolor{lightorange}\makebox[1.05cm]{2.57}
& \cellcolor{lightorange}\makebox[1.05cm]{1.76}
& \cellcolor{lightorange}\makebox[1.05cm]{0.58}
& \cellcolor{lightorange}\makebox[1.05cm]{6.07}
& \cellcolor{lightred}\makebox[1.05cm]{1.04}
& \cellcolor{lightred}\makebox[1.05cm]{1.36}
& \cellcolor{lightred}\makebox[1.05cm]{1.06}
& \cellcolor{lightred}\makebox[1.05cm]{0.35}
& \cellcolor{lightred}\makebox[1.05cm]{2.95}
\\

C$^5$ ~\cite{afifi2021cross}
  & \cellcolor{lightred}\makebox[1.05cm]{1.09} & \cellcolor{lightred}\makebox[1.05cm]{1.47}
  & \cellcolor{lightred}\makebox[1.05cm]{1.14}
  & \cellcolor{lightred}\makebox[1.05cm]{0.38}
  & \cellcolor{lightred}\makebox[1.05cm]{3.22}
  & \cellcolor{lightyellow}\makebox[1.05cm]{1.71}
  & \cellcolor{lightyellow}\makebox[1.05cm]{2.61}
  & \cellcolor{lightorange}\makebox[1.05cm]{1.85}
  & \cellcolor{lightyellow}\makebox[1.05cm]{0.63}
  & \cellcolor{lightorange}\makebox[1.05cm]{6.05} \\

PCC ~\cite{Wei2023ColorCF} & \makebox[1.05cm]{2.53} & \makebox[1.05cm]{3.91} & \makebox[1.05cm]{2.80} & \makebox[1.05cm]{0.86} & \makebox[1.05cm]{9.26} & \makebox[1.05cm]{3.57} & \makebox[1.05cm]{4.36} & \makebox[1.05cm]{3.71} & \makebox[1.05cm]{1.13} & \makebox[1.05cm]{8.90}
\\

ePCC ~\cite{liu2026color} 
& \makebox[1.05cm]{2.39} 
& \makebox[1.05cm]{3.52} 
& \makebox[1.05cm]{2.54} 
& \makebox[1.05cm]{0.86} 
& \makebox[1.05cm]{7.96} 
& \cellcolor{lightorange}\makebox[1.05cm]{1.64} 
& \cellcolor{lightorange}\makebox[1.05cm]{2.47} 
& \cellcolor{lightyellow}\makebox[1.05cm]{1.76} 
& \cellcolor{lightorange}\makebox[1.05cm]{0.58} 
& \makebox[1.05cm]{5.73}

\\

RL-AWB (Ours) & \cellcolor{lightyellow}\makebox[1.05cm]{1.98}
& \cellcolor{lightyellow}\makebox[1.05cm]{3.07}
& \cellcolor{lightyellow}\makebox[1.05cm]{2.24}
& \cellcolor{lightyellow}\makebox[1.05cm]{0.69}
& \cellcolor{lightyellow}\makebox[1.05cm]{7.22}
& \makebox[1.05cm]{3.01}
& \makebox[1.05cm]{3.22}
& \makebox[1.05cm]{3.03}
& \makebox[1.05cm]{1.43}
& \cellcolor{lightyellow}\makebox[1.05cm]{5.32} \\
\bottomrule
\end{tabular}%
}

\end{table}

\begin{table}[t]
\centering
\fontsize{8.5pt}{10pt}\selectfont
\caption{\textbf{In-dataset evaluation results on NCC and LEVI datasets.} Reproduction angular error in degrees. All learning-based baselines are implemented using three-fold cross-validation protocols and trained on the complete dataset; RL-AWB uses only 5 training images.}
\label{tab:sup_training_results}

\vspace{-5pt}

\resizebox{\textwidth}{!}{%
\begin{tabular}{p{2.6cm} | >{\centering\arraybackslash}p{0.95cm} >{\centering\arraybackslash}p{0.95cm} >{\centering\arraybackslash}p{0.95cm} >{\centering\arraybackslash}p{0.95cm} >{\centering\arraybackslash}p{0.95cm} | >{\centering\arraybackslash}p{0.95cm} >{\centering\arraybackslash}p{0.95cm} >{\centering\arraybackslash}p{0.95cm} >{\centering\arraybackslash}p{0.95cm} >{\centering\arraybackslash}p{0.95cm}}
\toprule
\multicolumn{1}{l|}{}
& \multicolumn{5}{c|}{NCC Dataset} & \multicolumn{5}{c}{LEVI Dataset} \\
\cmidrule(lr){2-6} \cmidrule(lr){7-11}
{Method} & Med. & Mean & Tri. & B-25 & W-25 & Med. & Mean & Tri. & B-25 & W-25 \\

\midrule

FFCC~\cite{barron2017fast}
& 3.89
& 5.02
& 4.76
& 1.89
& 13.7
& 5.10
& 8.12
& 5.29
& 2.35
& 18.7 \\

FC$^4$ ~\cite{hu2017fc4} 
& 14.5 
& 14.3 
& 14.3 
& 5.52 
& 22.9 
& 10.8 
& 11.7 
& 10.7 
& 4.52 
& 20.8 \\

$C^4$ ~\cite{yu2020cascading} 
& \cellcolor{lightred}2.22 
& \cellcolor{lightred}3.43 
& \cellcolor{lightred}2.48 
& \cellcolor{lightorange}0.82 
& \cellcolor{lightred}8.00 
& \cellcolor{lightorange}1.73 
& \cellcolor{lightred}2.20 
& \cellcolor{lightred}1.77 
& \cellcolor{lightorange}0.58 
& \cellcolor{lightred}4.74 \\

$C^5$ ~\cite{afifi2021cross} 
& \cellcolor{lightorange}2.36 
& \cellcolor{lightorange}3.56 
& \cellcolor{lightorange}2.60 
& \cellcolor{lightred}0.80 
& \cellcolor{lightorange}8.30 
& \cellcolor{lightred}1.68 
& \cellcolor{lightorange}2.30 
& \cellcolor{lightred}1.77 
& \cellcolor{lightred}0.54 
& \cellcolor{lightorange}5.20 \\

PCC ~\cite{Wei2023ColorCF} 
& 5.18 
& 6.01
& 6.28 
& 1.70 
& 15.1 
& 6.28 
& 7.37 
& 6.48 
& 2.23 
& 14.5 \\


ePCC ~\cite{liu2026color} 
& {3.12} 
& {4.66} 
& {3.46} 
& {1.08} 
& {10.51} 
& \cellcolor{lightyellow}{2.39} 
& \cellcolor{lightyellow}{3.61} 
& \cellcolor{lightorange}{2.66} 
& \cellcolor{lightyellow}{0.80} 
& \cellcolor{lightyellow}{8.45}
\\

RL-AWB (Ours) 
& \cellcolor{lightyellow}2.71 
& \cellcolor{lightyellow}4.13 
& \cellcolor{lightyellow}3.04 
& \cellcolor{lightyellow}0.97 
& \cellcolor{lightyellow}9.47 
& 5.07 
& 5.60 
& \cellcolor{lightyellow}5.15 
& 2.31 
& 9.80 \\

\bottomrule
\end{tabular}%
}

\normalsize
\end{table}

For learning-based approaches, C$^5$ achieves the best reproduction angular errors on both NCC and LEVI when trained with full data. Notably, our RL-AWB, despite using only 5 training images, achieves the third-best performance among learning-based methods. The combination of SGP-LRD and RL-AWB yields a strong and interpretable nighttime AWB solution when trained and evaluated within the same dataset. It is worth noting that GCC is a diffusion-based color constancy method that derives an illumination prior from well-illuminated training data. When deployed in nighttime low-light environments, the combination of pervasive noise and significant distribution shifts can compromise the stability of the illumination sampling process, resulting in extremely large estimation errors (90° angular error in the worst 25\% cases).

\subsection{Cross-dataset Generalization Comparison}

\cref{tab:sup_cross_dataset_vertical_tsi} summarizes the cross-dataset reproduction angular errors when training on one dataset and testing on the other. When trained on NCC and evaluated on LEVI, all fully-supervised learning-based methods suffer from substantial degradation, with median errors ranging from 10.30 (C$^5$) to 22.87 (GCC). In contrast, RL-AWB achieves a median of only 5.10 and attains the best performance across all reported statistics. The opposite direction, training on LEVI and testing on NCC, shows the same trend: RL-AWB consistently attains the lowest errors among all competing methods, with C$^5$ ranking second and GCC showing relatively competitive performance on this direction.

These results indicate that, despite being trained on a single dataset with only 5 images, RL-AWB generalizes well across sensors and scene distributions, providing stable nighttime white balance on unseen datasets and clearly outperforming fully supervised learning-based baselines.
\begin{table}[t]
\centering
\fontsize{8.5pt}{10pt}\selectfont
\caption{\textbf{Cross-dataset evaluation between NCC and LEVI datasets.} Reproduction angular error in degrees. All learning-based baselines are implemented using three-fold cross-validation protocols and trained on the complete dataset.}
\label{tab:sup_cross_dataset_vertical_tsi}

\vspace{-5pt}

\resizebox{\textwidth}{!}{%
\begin{tabular}{p{2.6cm} | >{\centering\arraybackslash}p{0.95cm} >{\centering\arraybackslash}p{0.95cm} >{\centering\arraybackslash}p{0.95cm} >{\centering\arraybackslash}p{0.95cm} >{\centering\arraybackslash}p{0.95cm} | >{\centering\arraybackslash}p{0.95cm} >{\centering\arraybackslash}p{0.95cm} >{\centering\arraybackslash}p{0.95cm} >{\centering\arraybackslash}p{0.95cm} >{\centering\arraybackslash}p{0.95cm}}

\toprule
\multicolumn{1}{c|}{}
& \multicolumn{5}{c|}{NCC $\rightarrow$ LEVI} & \multicolumn{5}{c}{LEVI $\rightarrow$ NCC} \\
\cmidrule(lr){2-6} \cmidrule(lr){7-11}
Method &
\makebox[1.05cm]{Med.} & \makebox[1.05cm]{Mean} & \makebox[1.05cm]{Tri.} & \makebox[1.05cm]{B-25} & \makebox[1.05cm]{W-25} & \makebox[1.05cm]{Med.} & \makebox[1.05cm]{Mean} & \makebox[1.05cm]{Tri.} & \makebox[1.05cm]{B-25} & \makebox[1.05cm]{W-25} \\
\midrule

FC$^4$ ~\cite{hu2017fc4}
& \makebox[1.05cm]{15.2} & \makebox[1.05cm]{15.9} & \makebox[1.05cm]{15.3} & \makebox[1.05cm]{11.3} & \cellcolor{lightyellow}\makebox[1.05cm]{22.0}
& \makebox[1.05cm]{16.2} & \makebox[1.05cm]{16.5} & \makebox[1.05cm]{16.2} & \makebox[1.05cm]{7.03} & \makebox[1.05cm]{27.0} \\

FFCC~\cite{barron2017fast}
&\cellcolor{lightorange}\makebox[1.05cm]{9.18}
&\cellcolor{lightorange}\makebox[1.05cm]{10.6}
&\cellcolor{lightorange}\makebox[1.05cm]{9.82}
&\cellcolor{lightorange}\makebox[1.05cm]{3.15}
&\makebox[1.05cm]{23.2}
&\makebox[1.05cm]{12.1}
&\makebox[1.05cm]{13.2}
&\makebox[1.05cm]{12.5}
&\makebox[1.05cm]{5.39}
&\makebox[1.05cm]{21.2} \\

$C^4$ ~\cite{yu2020cascading}
& \makebox[1.05cm]{15.1} & \makebox[1.05cm]{15.9} & \makebox[1.05cm]{15.2} & \makebox[1.05cm]{8.48} & \makebox[1.05cm]{24.6}
& \makebox[1.05cm]{13.8} & \makebox[1.05cm]{16.1} & \makebox[1.05cm]{14.8} & \makebox[1.05cm]{8.62} & \makebox[1.05cm]{26.0} \\
$C^5$ ~\cite{afifi2021cross}
& \cellcolor{lightyellow}\makebox[1.05cm]{10.3} & \makebox[1.05cm]{12.4} & \cellcolor{lightyellow}\makebox[1.05cm]{10.9} & \cellcolor{lightyellow}\makebox[1.05cm]{4.20} & \makebox[1.05cm]{24.3}
& \cellcolor{lightorange}\makebox[1.05cm]{5.91} & \cellcolor{lightorange}\makebox[1.05cm]{7.78} & \cellcolor{lightorange}\makebox[1.05cm]{6.51} & \cellcolor{lightorange}\makebox[1.05cm]{1.95} & \cellcolor{lightorange}\makebox[1.05cm]{16.4} \\
PCC ~\cite{Wei2023ColorCF}
& \makebox[1.05cm]{18.6} & \makebox[1.05cm]{19.7} & \makebox[1.05cm]{18.4} & \makebox[1.05cm]{10.8} & \makebox[1.05cm]{31.0}
& \makebox[1.05cm]{11.5} & \cellcolor{lightyellow}\makebox[1.05cm]{11.6} & \makebox[1.05cm]{11.2} & \makebox[1.05cm]{4.5} & \cellcolor{lightyellow}\makebox[1.05cm]{19.6} \\
GCC ~\cite{chang2025gcc}
& \makebox[1.05cm]{22.9} & \makebox[1.05cm]{26.0} & \makebox[1.05cm]{23.5} & \makebox[1.05cm]{12.4} & \makebox[1.05cm]{45.3}
& \makebox[1.05cm]{11.1} & \makebox[1.05cm]{11.7} & \cellcolor{lightyellow}\makebox[1.05cm]{10.8} & \cellcolor{lightyellow}\makebox[1.05cm]{3.10} & \makebox[1.05cm]{22.3} \\

ePCC ~\cite{liu2026color}
& \makebox[1.05cm]{11.1} & \cellcolor{lightyellow}\makebox[1.05cm]{11.8} & \makebox[1.05cm]{11.2} & \makebox[1.05cm]{8.0} & \cellcolor{lightorange}\makebox[1.05cm]{16.9} & 
\cellcolor{lightyellow}\makebox[1.05cm]{10.5} & \makebox[1.05cm]{11.9} & \makebox[1.05cm]{10.9} & \makebox[1.05cm]{6.5} & \makebox[1.05cm]{19.6} \\

RL-AWB (Ours)
& \cellcolor{lightred}\makebox[1.05cm]{5.10} & \cellcolor{lightred}\makebox[1.05cm]{5.62} & \cellcolor{lightred}\makebox[1.05cm]{5.19} & \cellcolor{lightred}\makebox[1.05cm]{2.35} & \cellcolor{lightred}\makebox[1.05cm]{9.80}
& \cellcolor{lightred}\makebox[1.05cm]{2.77} & \cellcolor{lightred}\makebox[1.05cm]{4.19} & \cellcolor{lightred}\makebox[1.05cm]{3.10} & \cellcolor{lightred}\makebox[1.05cm]{0.97} & \cellcolor{lightred}\makebox[1.05cm]{9.66} \\
\bottomrule
\end{tabular}%
}

\normalsize
\end{table}

\subsection{Runtime analysis.} Early termination reduces RL-AWB to 3 steps per image on average. Runtimes for C$^4$, FC$^4$, PCC, C$^5$, and GCC are 1.06, 1.2, 2.18, 7.41, and 145.3 ms, respectively, all GPU/CUDA-accelerated. In contrast, RL-AWB’s CPU runtime is dominated by SGP-LRD (87\% of the total latency). Since these per-pixel and local-statistics operations are inherently parallelizable, they can be readily integrated into ISP/GPU ASICs to achieve significant speedups. We leave this hardware optimization for future work.

\section{Additional Ablation Studies}
\label{sec:supablation}
\paragraph{Effect of model architecture.}
We study the impact of the backbone architecture by comparing single-branch and dual-branch designs under the same SAC configuration. As shown in \cref{tab:sup_ablation_arch}, the dual-branch variant consistently achieves lower errors. This is because our state comprises not only a high-dimensional RGB-uv histogram (10\,800 dimensions) but also a low-dimensional adjustment history (11 dimensions) encoding recent parameter values and the current step index. In a single-branch network, directly concatenating these two parts tends to dilute the influence of the low-dimensional signals. The dual-branch design, on the other hand, processes the histogram and history through separate MLPs to obtain two 64-dimensional embeddings, which are then concatenated and fused. This structure preserves the adjustment-related information more effectively, leading to better AWB parameter updates.
\begin{table}[t]
\centering
\fontsize{8.5pt}{10pt}\selectfont
\caption{\textbf{Ablation on network architecture} SAC algorithm, 5 training images.}
\label{tab:sup_ablation_arch}

\vspace{-5pt}

\begin{tabular}{>{\centering\arraybackslash}p{1cm} >{\centering\arraybackslash}p{0.95cm} >{\centering\arraybackslash}p{0.95cm} >{\centering\arraybackslash}p{0.95cm} >{\centering\arraybackslash}p{0.95cm} >{\centering\arraybackslash}p{0.95cm} >{\centering\arraybackslash}p{0.95cm}}

\toprule
\multirow{2}{*}{} & \multicolumn{3}{c}{NCC Dataset} & \multicolumn{3}{c}{LEVI Dataset} \\
\cmidrule(lr){2-4} \cmidrule(lr){5-7}
 & Med. & Mean & W-25 & Med. & Mean & W-25 \\

\midrule
Single      & 2.11 & 3.25 & 7.67 & 3.06 & 3.29 & 5.48 \\
\rowcolor{black!10}
\textbf{Dual} & \textbf{1.98} & \textbf{3.07} & \textbf{7.22} & \textbf{3.01} & \textbf{3.22} & \textbf{5.32} \\
\bottomrule
\end{tabular}
\end{table}

\begin{table}[t]
\centering
\fontsize{8.5pt}{10pt}\selectfont
\caption{\textbf{Ablation study on DRL algorithms} 5 training images.}
\label{tab:ablation_drl}

\vspace{-5pt}

\begin{tabular}{>{\centering\arraybackslash}p{1cm} >{\centering\arraybackslash}p{0.95cm} >{\centering\arraybackslash}p{0.95cm} >{\centering\arraybackslash}p{0.95cm} >{\centering\arraybackslash}p{0.95cm} >{\centering\arraybackslash}p{0.95cm} >{\centering\arraybackslash}p{0.95cm}}

\toprule
\multirow{2}{*}{} & \multicolumn{3}{c}{NCC Dataset} & \multicolumn{3}{c}{LEVI Dataset} \\
\cmidrule(lr){2-4} \cmidrule(lr){5-7}
 & Med. & Mean & W-25 & Med. & Mean & W-25 \\
 
\midrule
PPO        & 2.16 & 3.20 & 7.51 & 3.09 & 3.27 & 5.49 \\
\rowcolor{black!10}
\textbf{SAC} & \textbf{1.98} & \textbf{3.07} & \textbf{7.22} & \textbf{3.01} & \textbf{3.22} & \textbf{5.32} \\
\bottomrule
\end{tabular}
\end{table}

\paragraph{Effect of RL algorithm.}
We compare the off-policy SAC against the on-policy PPO under identical settings (\cref{tab:ablation_drl}). SAC consistently outperforms PPO on both datasets, benefiting from more sample-efficient off-policy updates and entropy-regularized exploration. The replay buffer in SAC enables better data reuse across the curriculum pool, whereas PPO's on-policy nature limits its ability to leverage past experience from different training images.


\section{Additional Visual Comparisons}
\label{sec:supvisual}
\begin{figure}[tp]
    \centering
    \includegraphics[width=\linewidth, trim={0cm 0cm 0cm 0cm},clip]{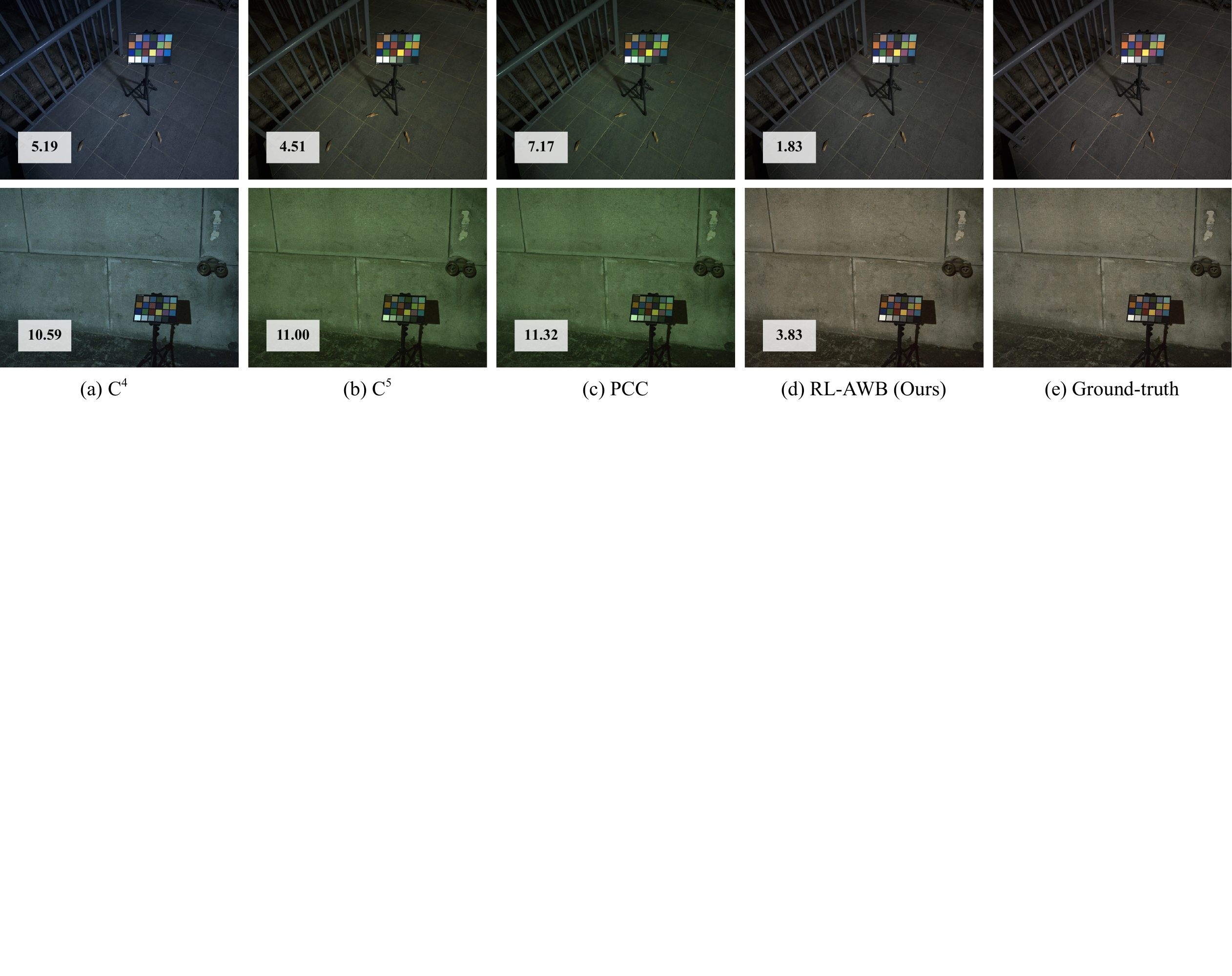}
    \caption{\textbf{Comparison of cross-sensor performance between our method and state-of-the-art methods on low-light nighttime images.}}
    \label{fig:WHYfigures_}
\end{figure}

\cref{fig:WHYfigures_} presents additional visual comparisons of cross-sensor performance between our method and state-of-the-art methods on low-light nighttime images. These results further demonstrate the superior color correction quality and cross-sensor robustness of RL-AWB compared to existing approaches.

\section{Future Work}
\label{sec:futurework}

First, the current agent controls only two AWB parameters, whereas the underlying SGP-LRD pipeline exposes multiple tunable parameters. Naively expanding the action space would substantially increase training complexity and cost. To address this, we plan to investigate structured and hierarchical policies, as well as low-dimensional latent action representations, to efficiently coordinate multiple ISP parameters. Second, while RL-AWB consistently reduces overall angular error, it may still over-correct a small number of challenging nighttime scenes, resulting in visually degraded outputs. Future work will therefore explore safety-aware reward formulations and constrained optimization strategies, such as penalizing abrupt parameter changes or incorporating preference-based regularization, to explicitly mitigate such failure cases. Third, the current implementation combines GPU-accelerated environment simulation with CPU-based reinforcement learning updates. Moving toward a fully GPU-resident training pipeline with batched rollouts could further reduce wall-clock training time and enable joint optimization across nighttime and daytime data, ultimately facilitating a unified all-time AWB agent.

\end{document}